\documentclass[conference]{IEEEtran}
\IEEEoverridecommandlockouts

\usepackage{cite}
\usepackage{amsmath,amssymb,amsfonts}
\usepackage{graphicx}
\usepackage{textcomp}
\usepackage{xcolor}

\usepackage{booktabs}
\usepackage{multirow}
\usepackage{array}

\usepackage[caption=false,font=footnotesize]{subfig}

\usepackage{tikz}
\usetikzlibrary{arrows.meta,shapes.geometric,positioning,fit,backgrounds,calc}

\usepackage[colorlinks=true,linkcolor=blue,citecolor=blue,urlcolor=blue,
            pdftitle={Enhancing Robotic Perception and Adaptability through Sensor Fusion and Origami-Inspired Designs},
            pdfauthor={Namai Chandra, Jaison Jose, Kavi Arya, Shivaram Kalyanakrishnan}]{hyperref}

\begin{document}

\title{Enhancing Robotic Perception and Adaptability through Sensor Fusion and Origami-Inspired Designs}

\author{
  \IEEEauthorblockN{Namai Chandra}
  \IEEEauthorblockA{\textit{IIT Madras}\\
    23f3000200@es.study.iitm.ac.in}
  \and
  \IEEEauthorblockN{Jaison Jose}
  \IEEEauthorblockA{\textit{e-Yantra, IIT Bombay}\\
    jaison@e-yantra.org}
  \and
  \IEEEauthorblockN{Kavi Arya}
  \IEEEauthorblockA{\textit{e-Yantra, IIT Bombay}\\
    kavi@cse.iitb.ac.in}
  \and
  \IEEEauthorblockN{Shivaram Kalyanakrishnan}
  \IEEEauthorblockA{\textit{e-Yantra, IIT Bombay}\\
    shivaram@cse.iitb.ac.in}
}

\maketitle

\begin{abstract}
Compact mobile robots must recover scene geometry under changing lighting and surface
texture while working within tight payload and cost limits.
We present a compact mobile robot that uses origami-inspired wheels for locomotion and
active control of its sensing geometry. As the wheels move between terrain-adaptive
configurations, the changing chassis pitch sweeps a 2D LiDAR through intermediate
elevations; held wheel
positions provide a chosen viewing angle. An IMU accounts for chassis attitude, and a
fusion node projects LiDAR returns into the RGB-D depth stream supplied to RTAB-Map.
The arrangement uses the wheel actuation already present on a sub-\$300, sub-2\,kg
prototype to extend the scanner's viewing geometry. We assess depth fusion in a
textureless indoor corridor and an outdoor sunlit area, with three runs per sensor
configuration in each setting. Mean full-frame invalid-depth fractions fell from
21\% to 11\% indoors and from 48\% to 18\% outdoors. The prototype combines improved
depth coverage with a continuously adjustable LiDAR viewpoint using the same actuation
that reconfigures its wheels.
\end{abstract}


%% =========================================================================
%%  System overview figure
%% =========================================================================
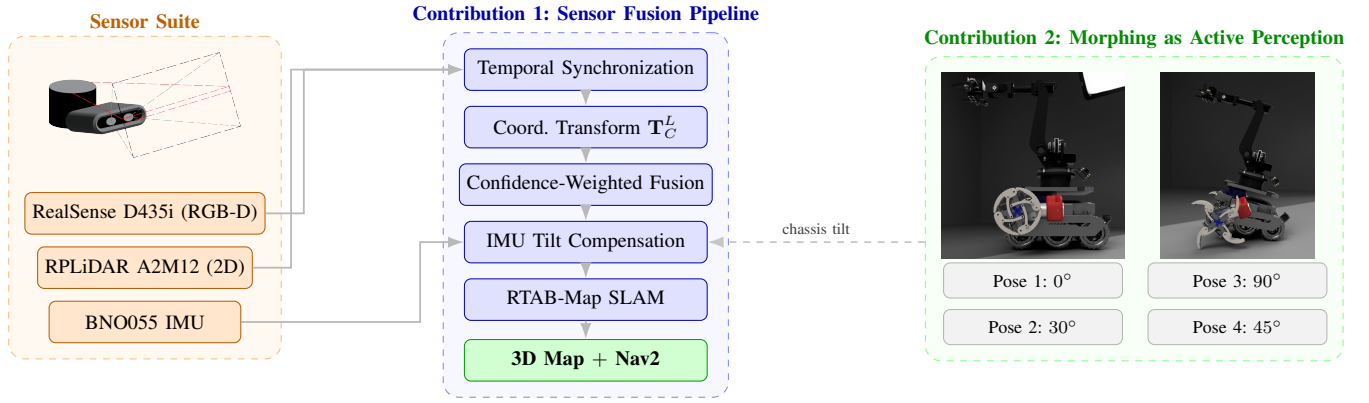
\begin{figure*}[!t]
\centering
\resizebox{0.98\textwidth}{!}{%
\begin{tikzpicture}[
  font=\small,
  snode/.style={draw=orange!70!black, rounded corners=3pt, fill=orange!20,
                minimum width=3.0cm, minimum height=0.65cm, align=center},
  pnode/.style={draw=blue!65!black, rounded corners=3pt, fill=blue!12,
                minimum width=3.8cm, minimum height=0.65cm, align=center},
  onode/.style={draw=green!60!black, rounded corners=3pt, fill=green!18,
                minimum width=3.8cm, minimum height=0.65cm, align=center,
                font=\small\bfseries},
  wnode/.style={draw=gray!60, rounded corners=3pt, fill=gray!10,
                minimum width=2.8cm, minimum height=0.55cm, align=center,
                font=\footnotesize},
  arr/.style={-{Latex[length=2.5mm,width=1.8mm]}, thick, draw=gray!55},
  darr/.style={-{Latex[length=2.5mm,width=1.8mm]}, thick, dashed, draw=gray!45},
]

%% COLUMN 1: SENSOR SUITE
\node[inner sep=0pt] (sfimg) at (2.1,-1.3)
  {\includegraphics[width=3.0cm]{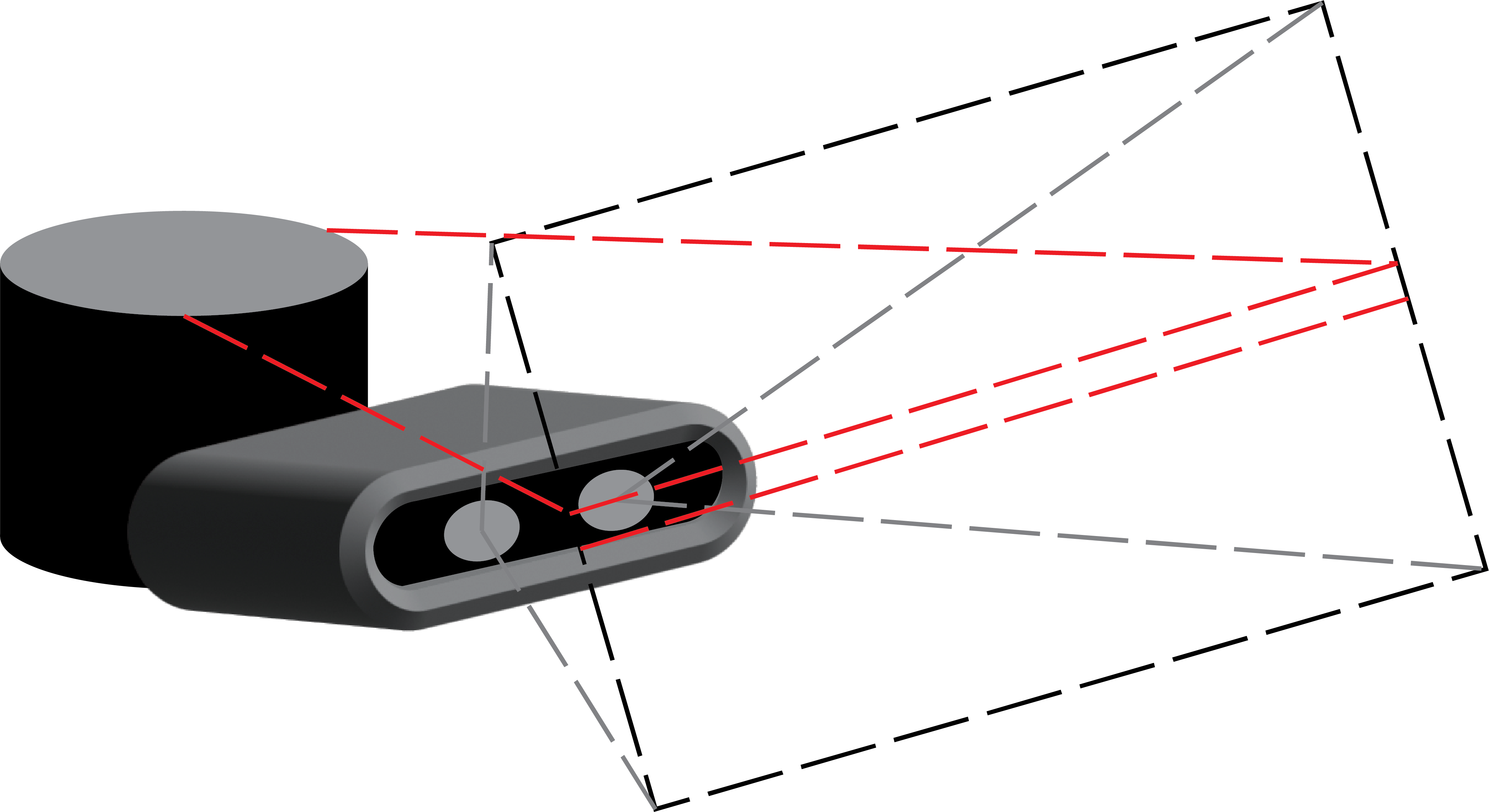}};

\node[snode] (dcam)   at (2.1,-3.0) {RealSense D435i (RGB-D)};
\node[snode] (lidar)  at (2.1,-3.85) {RPLiDAR A2M12 (2D)};
\node[snode] (imubox) at (2.1,-4.7)  {BNO055 IMU};

\begin{scope}[on background layer]
  \node[draw=orange!55, fill=orange!5, dashed, rounded corners=7pt,
        fit={(sfimg)(dcam)(lidar)(imubox)}, inner sep=7pt,
        label={[font=\small\bfseries, text=orange!75!black]above:Sensor Suite}]
        (col1box) {};
\end{scope}

%% COLUMN 2: PREPROCESSING PIPELINE
\node[pnode] (sync)      at (9.0,-0.75) {Temporal Synchronization};
\node[pnode] (transform) at (9.0,-1.65) {Coord. Transform $\mathbf{T}_{C}^{L}$};
\node[pnode] (fusion)    at (9.0,-2.55) {Confidence-Weighted Fusion};
\node[pnode] (tiltcomp)  at (9.0,-3.45) {IMU Tilt Compensation};
\node[pnode] (rtabmap)   at (9.0,-4.35) {RTAB-Map SLAM};
\node[onode] (mapout)    at (9.0,-5.30) {3D Map $+$ Nav2};

\begin{scope}[on background layer]
  \node[draw=blue!50, fill=blue!3, dashed, rounded corners=7pt,
        fit={(sync)(transform)(fusion)(tiltcomp)(rtabmap)(mapout)}, inner sep=7pt,
        label={[font=\small\bfseries, text=blue!70!black]above:%
          {\textbf{Contribution 1:} Sensor Fusion Pipeline}}]
        (col2box) {};
\end{scope}

%% COLUMN 3: MORPHING WHEEL PLATFORM
%% Enlarged CAD views sit side by side; labels denote selected milestones.
\node[inner sep=0pt] (poseA) at (16.0,-2.25)
  {\includegraphics[height=2.9cm]{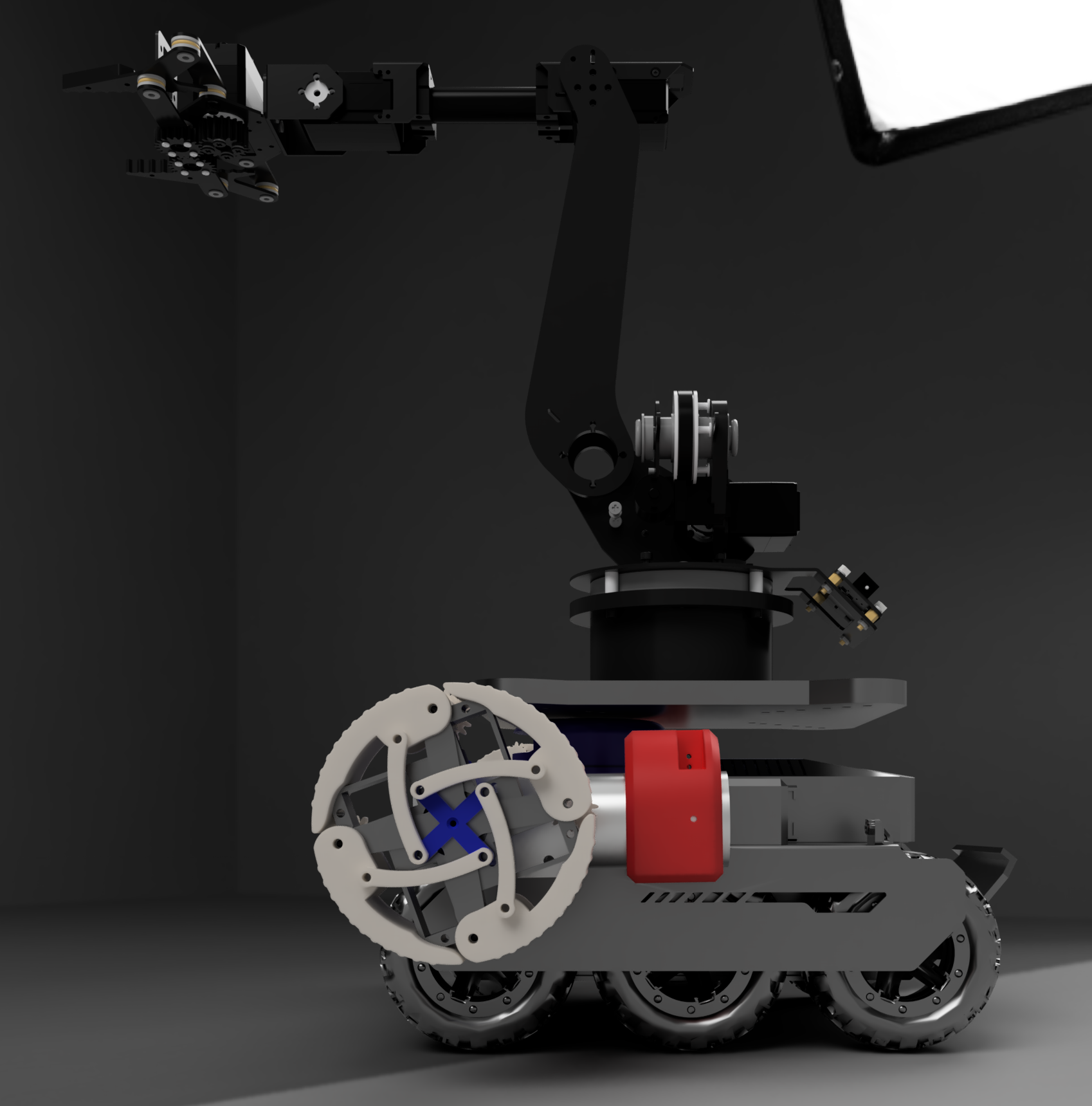}};
\node[inner sep=0pt] (poseD) at (19.2,-2.25)
  {\includegraphics[height=2.9cm]{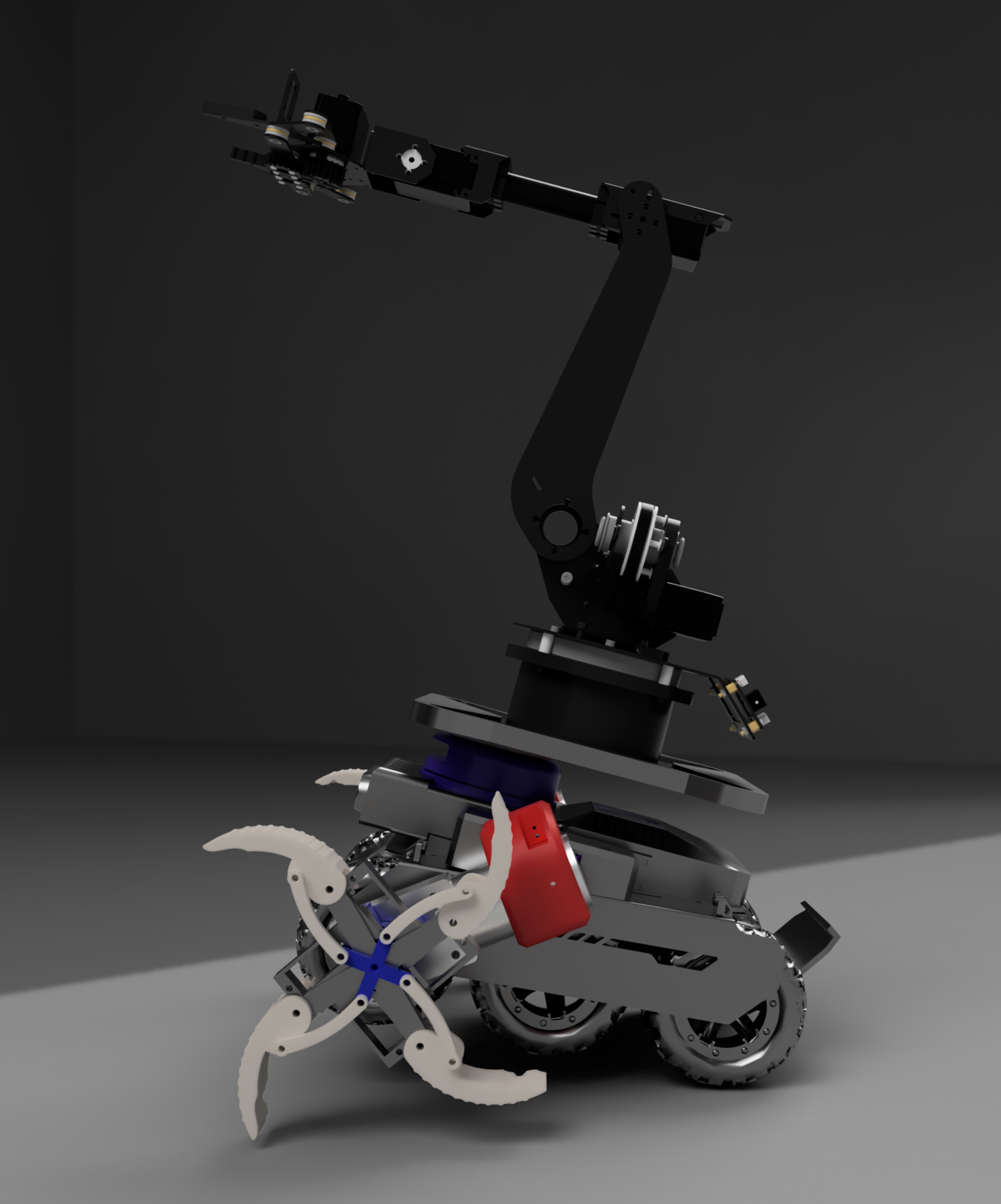}};
\node[wnode] (p1) at (16.0,-4.05) {Pose 1: $0^\circ$};
\node[wnode] (p3) at (19.2,-4.05) {Pose 3: $90^\circ$};
\node[wnode] (p2) at (16.0,-4.78) {Pose 2: $30^\circ$};
\node[wnode] (p4) at (19.2,-4.78) {Pose 4: $45^\circ$};

\begin{scope}[on background layer]
  \node[draw=green!50, fill=green!3, dashed, rounded corners=7pt,
        fit={(poseA)(p1)(p2)(p3)(p4)(poseD)}, inner sep=7pt,
        label={[font=\small\bfseries, text=green!55!black]above:%
          {\textbf{Contribution 2:} Morphing as Active Perception}}]
        (col3box) {};
\end{scope}

%% ARROWS: Sensors → Pipeline
\draw[arr] (dcam.east)   -- ++(0.6,0) |- (sync.west);
\draw[arr] (lidar.east)  -- ++(0.6,0) |- (sync.west);
%% IMU arrow: long horizontal run before turning up; keeps the line clear of
%% the camera/LiDAR arrows.
\draw[arr] (imubox.east) -- ++(2.8,0) |- (tiltcomp.west);

%% ARROWS: Pipeline internal
\draw[arr] (sync)      -- (transform);
\draw[arr] (transform) -- (fusion);
\draw[arr] (fusion)    -- (tiltcomp);
\draw[arr] (tiltcomp)  -- (rtabmap);
\draw[arr] (rtabmap)   -- (mapout);

%% ARROWS: Morphing platform → Pipeline (chassis tilt feedback)
\draw[darr] (col3box.west |- tiltcomp.east)
  -- node[above, font=\scriptsize, text=black!75] {chassis tilt}
  (tiltcomp.east);

\end{tikzpicture}%
}% end resizebox
\caption{System overview.
  \textbf{Left:} sensor suite combining a RealSense D435i RGB-D camera, an RPLiDAR
  A2M12 2D scanner, and a BNO055 IMU.
  \textbf{Center (Contribution 1):} confidence-weighted fusion pipeline that
  blends LiDAR depth into the RGB-D stream before passing the result to RTAB-Map.
  \textbf{Right (Contribution 2):} origami morphing wheels tilt the chassis to
  sweep the 2D LiDAR vertically; the IMU compensates the tilt in software. The
  flat (left) and fully unfolded (right) views are CAD renders. The four labelled
  poses are milestones along a continuous wheel motion, during which the LiDAR
  sweeps through intermediate elevations.}
\label{fig:hero}
\end{figure*}
%% =========================================================================

\section{Introduction}\label{sec:intro}

Compact mobile robots are of growing interest for constrained search-and-rescue (SAR)
scenarios such as collapsed buildings, rubble-strewn corridors, and partially obstructed
indoor spaces \cite{murphy2004,darpasubt2022}. Initiatives like DARPA SubT have highlighted
the appeal of small, reconfigurable platforms in such settings. A persistent challenge,
however, is achieving usable 3D environmental perception on a robot that is also small,
lightweight, and inexpensive enough to be considered semi-expendable. Our system
addresses these constraints with an origami-inspired chassis whose wheel poses also
steer a 2D LiDAR's scan plane. We study the resulting fusion and active scanning on a
compact prototype in indoor and outdoor sunlit tests.

\subsection{Depth Sensing on a Small Robot}
No single low-cost sensor provides both accurate range and visual detail in every
setting.

\textbf{Stereo vision} estimates depth by triangulating the disparity between two
viewpoints. While intuitive, its accuracy degrades rapidly with distance (depth error grows
quadratically with range), and it fails entirely on textureless surfaces such as blank
walls, uniform rubble, or featureless corridors, where no visual features exist to match.

\textbf{Active-stereo depth cameras}, such as the Intel RealSense D435i, use two IR
imagers and an optional projected texture to improve stereo matching \cite{b8}.
Projected texture helps on low-texture surfaces nearby, while range and direct sunlight
can reduce depth completeness. The D435i provides both color images and a per-pixel depth
map for appearance-based visual SLAM and geometric verification.

\textbf{LiDAR sensors} measure distance by timing laser pulse returns, providing
millimeter-to-centimeter accuracy invariant to lighting conditions and surface texture.
A 3D LiDAR (e.g., Velodyne VLP-16, Ouster OS1) produces dense omnidirectional point
clouds, but these units cost \$1{,}000--\$10{,}000+ and weigh 500\,g--1\,kg, impractical
for small, expendable disaster-response robots. A 2D LiDAR such as the RPLiDAR A2M12 is
lightweight ($\sim$190\,g), affordable ($\sim$\$100), and accurate up to 12\,m \cite{7},
but captures depth in only a single horizontal plane with no associated visual
information, insufficient for full 3D mapping or appearance-based loop closure.

We combine the camera's dense image with LiDAR range where their fields of view
overlap. Wheel reconfiguration then changes the LiDAR scan plane as the robot moves
through its operating geometry.

\begin{figure}[!htbp]
  \centering
  \includegraphics[width=2.4in]{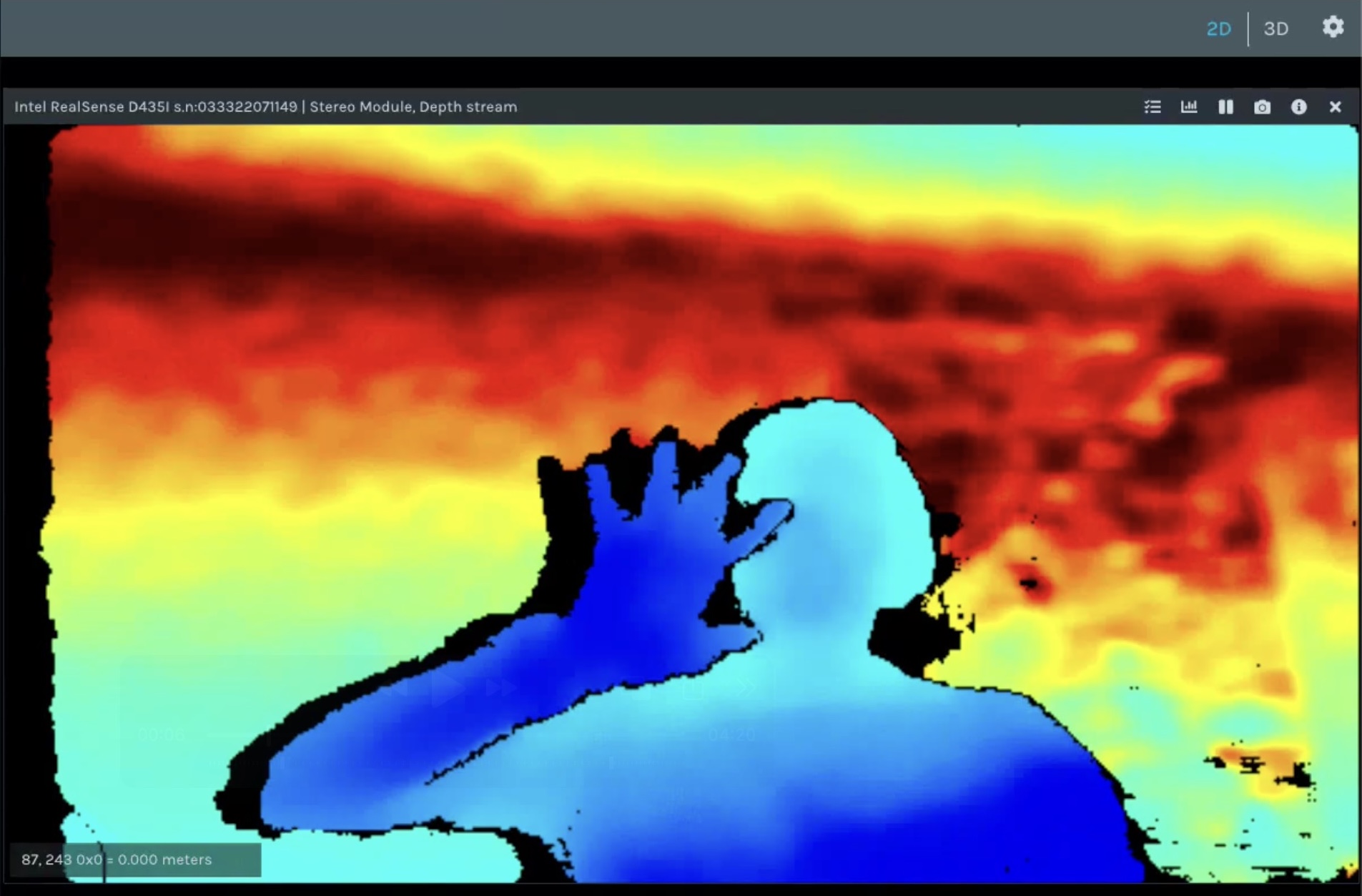}
  \caption{Raw depth image from Intel RealSense D435i. Black regions indicate depth voids
    where the IR pattern is too sparse (beyond $\sim$3\,m) or where surface texture is
    insufficient for stereo matching.}
  \label{fig:depth_image}
\end{figure}

\subsection{How Fusion and Morphing Are Combined}

Our design couples RGB-D and 2D LiDAR fusion with the robot's origami-inspired chassis.
Wheel reconfiguration changes the robot's geometry for traversal and simultaneously
steers the LiDAR scan plane:

\begin{enumerate}
  \item \textbf{Fusion mitigates the RGB-D camera's intrinsic weaknesses.}
    Invalid or low-confidence stereo depth pixels are augmented by a confidence-weighted
    blend with 2D LiDAR returns within the camera--LiDAR overlap. In our test runs this
    reduced per-frame depth-void
    fractions in both the indoor textureless and outdoor sunlit scenarios
    (Section~\ref{sec:results}); we report this as a descriptive statistic of the depth
    stream, not as an accuracy benchmark.

  \item \textbf{The morphing wheels are re-used to drive a tilted LiDAR sweep.}
    A 2D LiDAR on a flat, fixed-wheel chassis is confined to a single scan plane;
    dedicated ``nodding'' servo mechanisms have long been used to pitch a 2D scanner
    \cite{nodding_lidar,wulf2003}. Rather than add a dedicated nodding servo, we re-use
    the locomotive degree of freedom provided by the morphing wheels to tilt the chassis,
    for active perception \cite{bajcsy1988}. During a wheel transition, the LiDAR
    samples intermediate elevations without a separate scanner servo
    (Section~\ref{sec:results-vertical}).

  \item \textbf{IMU-based attitude estimation prevents the tilt from corrupting the map.}
    A Madgwick gradient-descent filter estimates chassis attitude from the onboard IMU,
    so that each tilted LiDAR scan is projected into a gravity-aligned global frame.
    Without this step the morphing-induced tilt would inject systematic error into the
    map rather than add vertical coverage.
\end{enumerate}

The full sensor stack is under \$300 and under 2\,kg on a custom prototype platform.
The perception unit can be detached and tripod-mounted for handheld mapping, which is
how parts of the dataset reported here were collected.

\subsection{Scope and Positioning}\label{sec:scope}

Prior work establishes RGB-D and LiDAR depth fusion
\cite{4,5,new_fusion1,new_fusion2,rgbd_lidar_fusion_prior}, mechanical sweeping of a
2D scanner \cite{nodding_lidar,wulf2003,morales2021}, and morphing wheel structures
\cite{felton2014,adaptive_morphology,b9,2}. We connect these capabilities on one compact robot: wheel movement changes the
LiDAR viewpoint, the IMU aligns scans during reconfiguration, and the fusion node
supplies depth to a standard SLAM backend. We measure depth completeness under two
sensor stressors and document the scan motion produced by the wheels. Table~\ref{tab:novelty}
locates the design choices relative to the relevant prior work.

\begin{table}[!t]
  \caption{How established techniques are combined in the prototype.}
  \label{tab:novelty}
  \centering
  \renewcommand{\arraystretch}{1.2}
  \footnotesize
  \begin{tabular}{@{}>{\raggedright\arraybackslash}p{0.27\columnwidth}
                  >{\raggedright\arraybackslash}p{0.17\columnwidth}
                  >{\raggedright\arraybackslash}p{0.45\columnwidth}@{}}
    \toprule
    \textbf{Component} & \textbf{Prior work} & \textbf{Implementation here} \\
    \midrule
    Pixel-level RGB-D + 2D LiDAR depth fusion
      & \cite{4,5,new_fusion1,new_fusion2,rgbd_lidar_fusion_prior}
      & Re-implemented as an upstream ROS~2 node feeding an unmodified SLAM backend. \\
    Pseudo-3D scanning by tilting a 2D LiDAR
      & \cite{nodding_lidar,wulf2003,morales2021}
      & Driven by the existing locomotive DOF (chassis tilt) instead of a dedicated nodding servo. \\
    Origami / flasher-pattern adaptive wheels
      & \cite{felton2014,adaptive_morphology,b9,2}
      & Re-used as both traversal mechanism and tilt actuator for the pseudo-3D scan. \\
    \bottomrule
  \end{tabular}
\end{table}

\section{Related Work}\label{sec:related}

\subsection{RGB-D and LiDAR Fusion}

Geometric alignment of camera and LiDAR measurements is well established. Dieterle
et al.~\cite{4} combine RGB-D and LiDAR for object tracking, and Song et al.~\cite{5}
use both sensors for relative localization. Zheng et al.~\cite{rgbd_lidar_fusion_prior}
study indoor 3D mapping with a 2D LiDAR and RGB-D camera; Feng and Kuang
\cite{new_fusion1} and Liu et al.~\cite{new_fusion2} also combine these sensing modes
for mapping or navigation. Wu et al.~\cite{b2} review learned LiDAR fusion methods.
We use a lightweight geometric fusion node because our immediate goal is to supply a
more complete depth image to RTAB-Map on the Jetson Orin Nano.

\subsection{Active Scanning and Mapping}

Surmann et al.~\cite{nodding_lidar} and Wulf et al.~\cite{wulf2003} sweep a 2D LiDAR
with a dedicated actuator to build 3D measurements. Our wheel mechanism also changes
the scan-plane elevation, but its actuator serves locomotion and cannot position the
scanner independently of the chassis. Morales et al.~\cite{morales2021} examine 2D
LiDAR and RGB-D integration on a mobile robot. Our design joins that sensor pairing
to chassis-driven scanning.

We use RTAB-Map's existing graph-based backend and visual loop closure
\cite{7}. Other visual or visual-inertial systems include ORB-SLAM3
\cite{orbslam3} and VINS-Fusion \cite{vinsfusion}; LiDAR-centred alternatives include
Cartographer \cite{cartographer}, LOAM \cite{loam}, LIO-SAM \cite{liosam}, and
FAST-LIO2 \cite{fastlio2}. We leave RTAB-Map unchanged and improve the depth stream
it receives where camera and LiDAR observations overlap.

\subsection{Origami-Inspired Wheels}

Origami mechanisms provide compact structures that can change shape with few inputs
\cite{rustolley2015,felton2014,adaptive_morphology,2,b4}. The flasher pattern
\cite{b9} motivates our folding wheel geometry. We use wheel reconfiguration for
terrain adaptation and to change the LiDAR viewpoint as the chassis tilts. The
connection between movement and sensing follows the active-perception principle of
choosing sensor viewpoints through action \cite{bajcsy1988,bohg2017}. Mechanical
scanning itself is established \cite{nodding_lidar,wulf2003}; here the wheel motion
supplies that scan motion on the same prototype.

\section{Background}\label{sec:background}

\subsection{Coordinate Frames and Projection}

We use rigid-body transformations in $SE(3)$ to align the sensor measurements. The
transformation $\mathbf{T} \in SE(3)$ between
the LiDAR frame $\{L\}$ and the Camera frame $\{C\}$ as a homogeneous $4 \times 4$ matrix:
\begin{equation}
  \mathbf{T}_{C}^{L} = \begin{bmatrix}
    \mathbf{R}_{3\times3} & \mathbf{t}_{3\times1} \\
    \mathbf{0}^T & 1
  \end{bmatrix}
  \label{eq:transform}
\end{equation}
where $\mathbf{R}$ represents the rotation matrix derived from the mounting orientation of
the RPLiDAR, and $\mathbf{t}$ represents the physical translation vector $[x,y,z]^T$.

To fuse 2D LiDAR scans with 3D depth maps, we employ the standard pinhole camera model.
A 3D point $\mathbf{p}_c = [x_c, y_c, z_c]^T$ in the camera frame is projected onto the
image plane $(u,v)$ via:
\begin{equation}
  u = \frac{f_x \cdot x_c}{z_c} + c_x, \quad
  v = \frac{f_y \cdot y_c}{z_c} + c_y
  \label{eq:projection}
\end{equation}
where $f_x, f_y$ are the focal lengths and $c_x, c_y$ are the principal points, obtained
from the RealSense factory intrinsic calibration \cite{zhang2000}. This projection is
critical for the depth-inpainting step used to fill voids in the stereoscopic depth map
with precise LiDAR returns.

\section{Methodology}\label{sec:methodology}

The architecture is organized as a multi-stage pipeline: raw sensor data flows
through fusion and tilt compensation before reaching the SLAM backend for map
construction. Hardware and software settings are given after the method in
Section~\ref{sec:implementation}.

\subsection{Perception Module}\label{sec:perception}

The perception module projects 2D LiDAR returns into the RGB-D camera frame, fuses depth in their overlapping field of view, and supplies the resulting stream to RTAB-Map. The following stages define the method.

\subsubsection{Preprocessing Pipeline}

The preprocessing pipeline is central to the system: it re-implements confidence-weighted
RGB-D and 2D LiDAR depth fusion (an established idea; see Section~\ref{sec:scope}) as an
upstream ROS~2 node so that an unmodified SLAM backend receives a cleaner depth stream. It
operates as follows before any data reaches RTAB-Map:

\begin{enumerate}
  \item \textbf{Raw Data Acquisition:} The RealSense publishes color images and depth
    images at 30\,Hz. The RPLiDAR publishes $360^\circ$ scans at 10\,Hz.
  \item \textbf{Temporal Synchronization:} The most recent LiDAR scan is paired with each
    incoming depth image using approximate time synchronization (tolerance: 50\,ms).
  \item \textbf{Coordinate Transform:} Each LiDAR point is transformed from the LiDAR
    frame to the camera frame using the extrinsic calibration matrix $\mathbf{T}_{C}^{L}$
    (detailed in Section~\ref{sec:methodology}).
  \item \textbf{Camera Projection:} Transformed LiDAR points are projected onto the
    camera's image plane using the pinhole model (\ref{eq:projection}) with the D435i's
    factory-calibrated intrinsics \cite{zhang2000}.
  \item \textbf{Confidence-Weighted Depth Fusion:} For each projected LiDAR pixel
    $(u,v)$, the node assigns a per-sensor confidence and forms a \emph{weighted blend}
    rather than a hard replacement. The camera is given a base confidence of $1.0$,
    scaled down by a factor of $0.7$ for far returns ($d_\mathrm{cam} > 5.0$\,m); the
    LiDAR is given a base confidence of $0.8$, halved for very close returns
    ($d_\mathrm{lidar} < 0.5$\,m). Wherever the LiDAR confidence exceeds a threshold
    $\tau_c = 0.6$ and both depths are valid, the fused depth is the weighted average
    \begin{equation}
      d_\mathrm{fused}(u,v) = (1-w_L)\,d_\mathrm{cam}(u,v) + w_L\,d_\mathrm{lidar}(u,v),
      \label{eq:fusion_blend}
    \end{equation}
    with LiDAR weight $w_L = 0.7$. When camera depth is invalid and a valid LiDAR return
    projects to that pixel, the node uses the LiDAR depth alone. The original camera
    value is retained outside the projected overlap or when no valid LiDAR return is
    available. This is related to the depth-completion problem \cite{ku2018,matoyou},
    where sparse range measurements densify a depth map, but is implemented here as a
    lightweight real-time preprocessing node rather than a learning-based postprocess. A
    stricter camera--LiDAR consistency gate (rejecting blends where
    $|d_\mathrm{cam}-d_\mathrm{lidar}|$ exceeds a fixed bound) is a natural extension that
    we use below only as an offline diagnostic (Section~\ref{sec:metric}), not as a live
    fusion gate.
  \item \textbf{Enhanced Depth Publication:} The fused depth image is published on a new
    ROS topic. The original, unmodified color image is published alongside it.
  \item \textbf{RTAB-Map Ingestion:} RTAB-Map subscribes to the \emph{fused} depth topic
    and the \emph{original} color topic. From RTAB-Map's perspective, it receives a standard
    RGB-D stream with fewer depth voids in the tested settings.
\end{enumerate}

This pipeline design means RTAB-Map operates on pre-cleaned data without any internal
modification to its codebase; we improved the quality of its input.

\subsubsection{FOV and Data Utilization}

The LiDAR's $360^\circ$ scan is divided into two functional streams. For depth fusion with the
camera, the LiDAR scan is cropped to an $80^\circ$ sector ($\pm 40^\circ$ about the
optical axis, parameter \texttt{fov\_angle}) approximately matching the D435i's horizontal
field of view ($87^\circ \pm 3^\circ$, reduced to account for edge distortion). Only within
this overlapping region are LiDAR depth values projected onto the camera's image plane. The
remaining $\sim 280^\circ$ of LiDAR data is not discarded: it is retained and published as a
standard \texttt{/scan} topic to the Nav2 navigation stack, providing $360^\circ$ obstacle
detection for path planning and obstacle avoidance.

\begin{figure}[!t]
  \centering
  \includegraphics[width=3.3in]{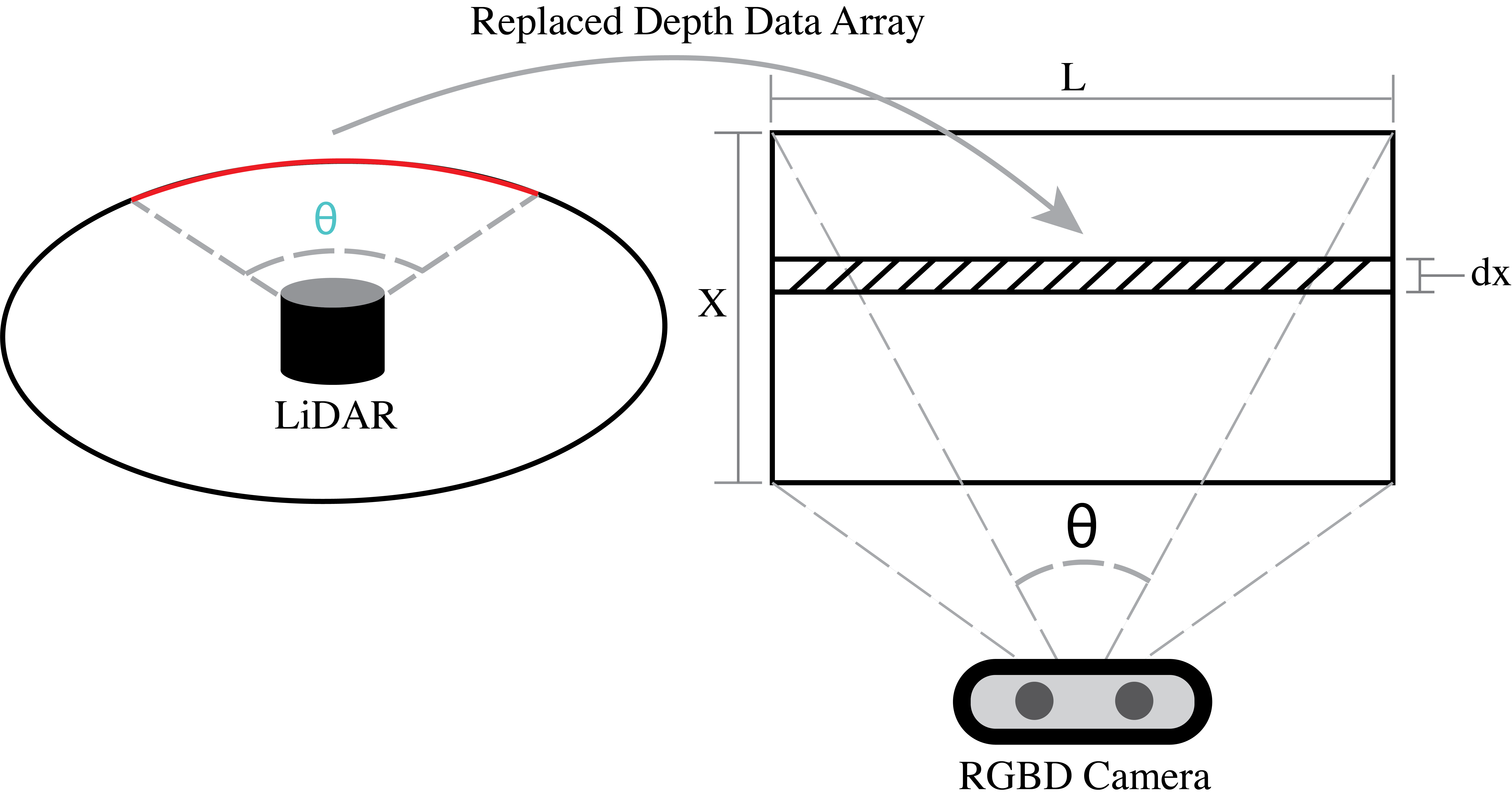}
  \caption{Sensor fusion in the camera--LiDAR overlap sector: LiDAR returns are blended
    with invalid or low-confidence stereo depth pixels (confidence-gated weighted average,
    $w_L=0.7$), producing a more complete depth image for RTAB-Map.}
  \label{fig:data_replacement}
\end{figure}

The alignment between sensors is achieved through the extrinsic transformation matrix
$\mathbf{T}_{C}^{L}$. The LiDAR is mounted 15\,mm above and 70\,mm forward of the RealSense
camera along the optical axis, with a rotation about the vertical ($z$) axis that aligns
the LiDAR's zero-angle with the camera's forward direction. The deployed system applies the
CAD-calibrated mounting angle \texttt{lidar\_rotation\_deg}~$=261^\circ$ (i.e., the nominal
$-90^\circ$ sensor rotation plus a small measured mounting correction); for clarity we
illustrate the transform below at the nominal $\theta = -90^\circ$. The translation offsets
($t_z = 15$\,mm, $t_x = -70$\,mm) were measured mechanically from the 3D-printed mount
geometry and verified against manufacturer CAD models:
\[
  \mathbf{T}_{C}^{L} = \begin{bmatrix}
    \cos(\theta) & -\sin(\theta) & 0 & t_x \\
    \sin(\theta) &  \cos(\theta) & 0 & t_y \\
    0 & 0 & 1 & t_z \\
    0 & 0 & 0 & 1
  \end{bmatrix}
\]
where $\theta = -90^\circ$, $t_x = -0.070$\,m, $t_y = 0$\,m, $t_z = 0.015$\,m.
Evaluating:
\[
  \mathbf{T}_{C}^{L} = \begin{bmatrix}
    0 & 1 & 0 & -0.070 \\
    -1 & 0 & 0 & 0 \\
    0 & 0 & 1 & 0.015 \\
    0 & 0 & 0 & 1
  \end{bmatrix}
\]
This matrix transforms LiDAR points $\mathbf{P}_L = [x_L,y_L,z_L,1]^T$ to camera
coordinates $\mathbf{P}_C = \mathbf{T}_{C}^{L}\mathbf{P}_L$, enabling precise spatial
correspondence between the two sensors \cite{4,5}.

\subsubsection{Computational Benefit of Pre-Fusion}

A key architectural decision was to fuse sensor data \emph{before} publishing to RTAB-Map,
rather than running parallel visual and LiDAR odometry threads within it. This
single-pipeline approach reduces the computational load on the resource-constrained Jetson
Orin Nano: instead of running two odometry estimators and a late-fusion module, the system
runs one fused depth stream through a single visual odometry pipeline, driven by practical
necessity given the 8\,GB shared GPU/CPU memory constraint.

\begin{figure}[!t]
  \centering
  \includegraphics[width=2.7in]{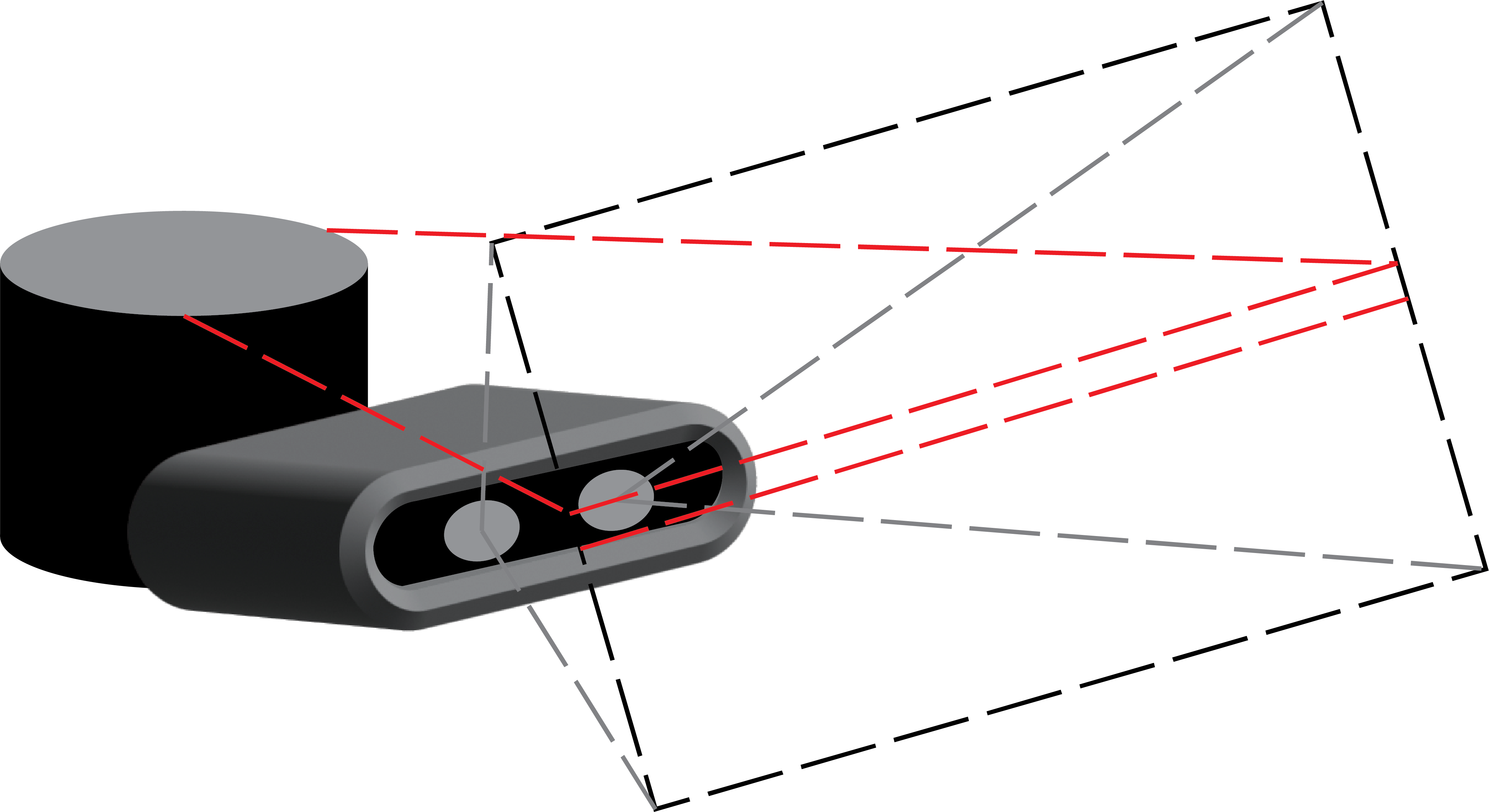}
  \caption{System pipeline: raw sensor data flows through the fusion preprocessing node
    (our contribution) before reaching RTAB-Map (existing library) for final map
    construction.}
  \label{fig:sensor_fusion_diagram}
\end{figure}

\subsection{Graph-Based SLAM Backend}

We pass the fused depth image and the original color image to RTAB-Map without
changing its odometry, loop-closure, or graph-optimization code. RTAB-Map represents
the trajectory as poses $\mathbf{x}=\{x_1,\ldots,x_T\}$ and optimizes the graph
constraints \cite{7,g2o}:
\begin{equation}
  \mathbf{x}^* = \underset{\mathbf{x}}{\operatorname{argmin}}
    \sum_{\langle i,j \rangle \in \mathcal{C}}
      \mathbf{e}_{ij}^T \boldsymbol{\Omega}_{ij}\, \mathbf{e}_{ij}.
  \label{eq:optimization}
\end{equation}
Here $\mathbf{e}_{ij}$ is the pose-constraint error and $\boldsymbol{\Omega}_{ij}$
is its information matrix. Our fusion node changes the depth measurements available
for visual odometry and geometric verification of candidate loop closures in the
camera--LiDAR overlap. The indoor loop-closure counts in Section~\ref{sec:results-slam}
show the effect observed in our recordings; we do not infer a change in the optimizer's
convergence or absolute trajectory accuracy from those counts.

\begin{figure}[!t]
  \centering
  \includegraphics[clip,trim={1230bp 0 0 0},width=3.3in]{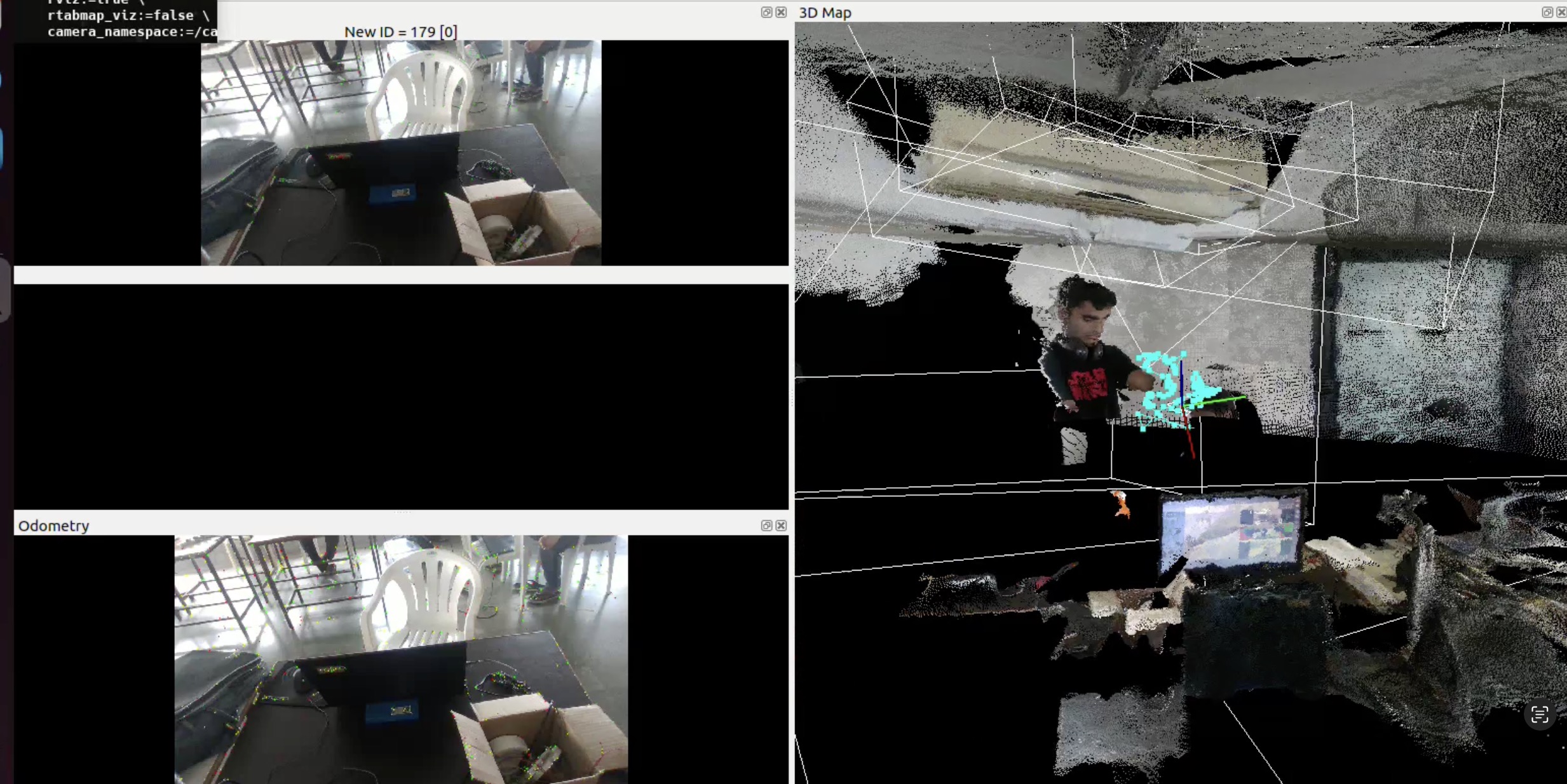}
  \caption{Reconstructed 3D point cloud generated by RTAB-Map using the fused depth
    stream (cropped from the RTAB-Map GUI). Regions that appear as voids in the
    camera-only raw depth are filled in by the LiDAR-augmented stream, producing a
    continuous representation of the scene.}
  \label{fig:3d_map}
\end{figure}

\subsection{Origami-Inspired Morphing Wheels}\label{sec:wheels}

The robot's morphing wheels utilize a flasher design, a radial folding pattern that
enables significant shape transformation through single-degree-of-freedom actuation
\cite{b9}. Developed by Jeremy Shafer, this compliant mechanism allows the chassis to
transition between functional geometries.

\subsubsection{Mechanism Description}

Each wheel module has four flaps ($n=4$) extending from a central hub and is fabricated
from 3D-printed PLA. Four servos per wheel, each rated at $\tau_s=2.2$\,kg$\cdot$cm,
drive a spur gear train with ratio $\eta=1.3$. The train engages a ring gear on the
inner surface of the flasher pattern's base ring and moves the flaps together. As the
flaps follow their pre-scored mountain and valley folds, the wheel changes from a
compact disk to an extended profile. Servo coordination controls the shape during
the transition.

\begin{figure}[!t]
  \centering
  \subfloat[Open configuration (extended state).\label{fig:w_open}]{%
    \includegraphics[width=0.46\columnwidth]{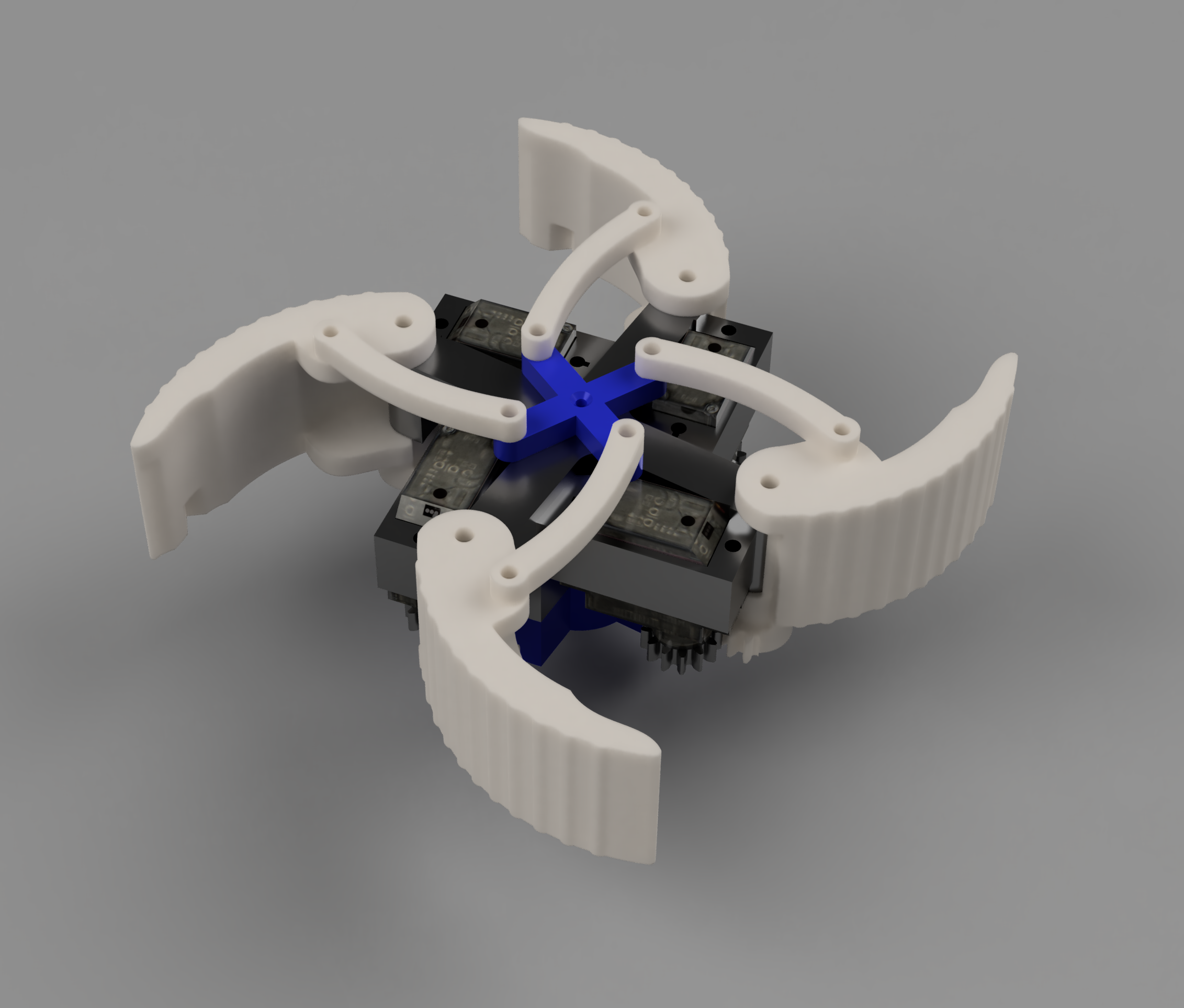}}
  \hfill
  \subfloat[Closed (morphed) configuration.\label{fig:w_close}]{%
    \includegraphics[width=0.46\columnwidth]{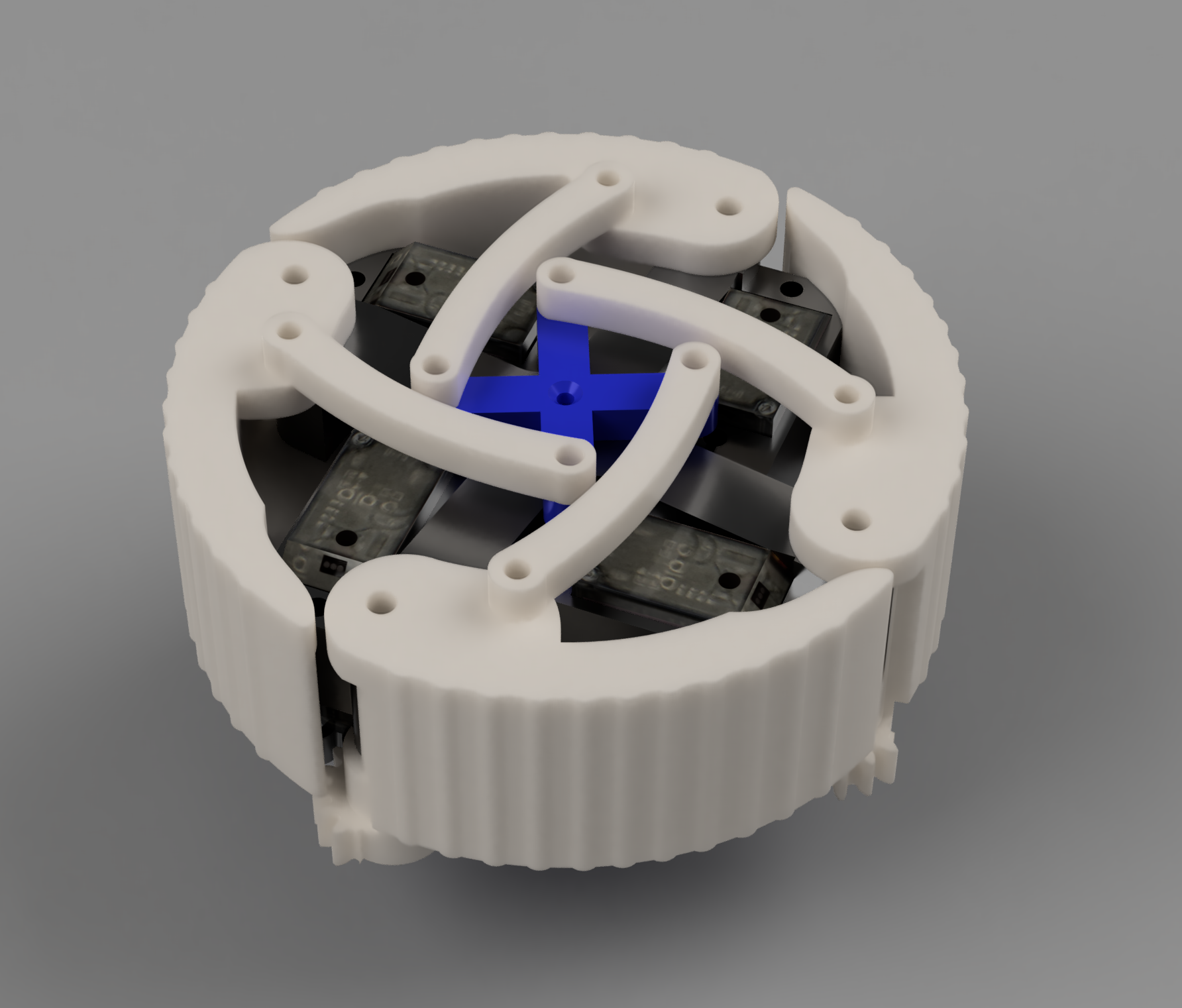}}
  \caption{CAD-rendered views of the origami morphing wheel in its two primary configurations.}
  \label{fig:wheels_combined}
\end{figure}

The open and closed wheel configurations are shown in Fig.~\ref{fig:wheels_combined}.

\subsubsection{Kinematic and Static Analysis}

\begin{figure}[!t]
  \centering
  \subfloat[Pose 1 (Baseline): wheels at $0^\circ$, flat chassis.\label{fig:pose1}]{%
    \includegraphics[width=0.46\columnwidth]{pose1.png}}
  \hfill
  \subfloat[Pose 2 (Tilted): $30^\circ$ for ground traction.\label{fig:pose2}]{%
    \includegraphics[width=0.46\columnwidth]{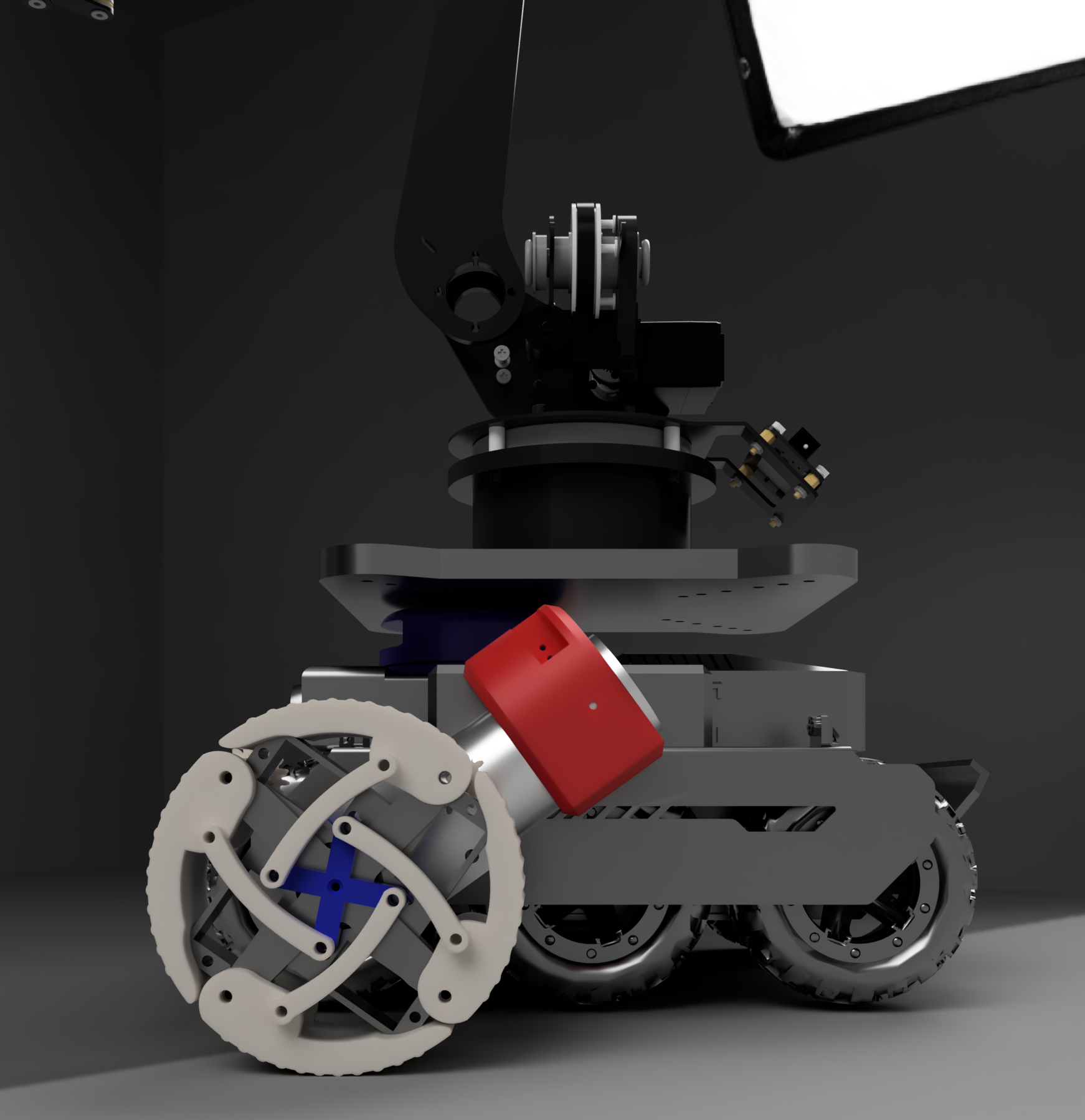}}\\[2pt]
  \subfloat[Pose 3 (Unfolded): $90^\circ$ for rough terrain and elevated LiDAR sweep.\label{fig:pose3}]{%
    \includegraphics[width=0.46\columnwidth]{pose3.png}}
  \hfill
  \subfloat[Pose 4 (Raised): $45^\circ$ for horizon adjustment.\label{fig:pose4}]{%
    \includegraphics[width=0.46\columnwidth]{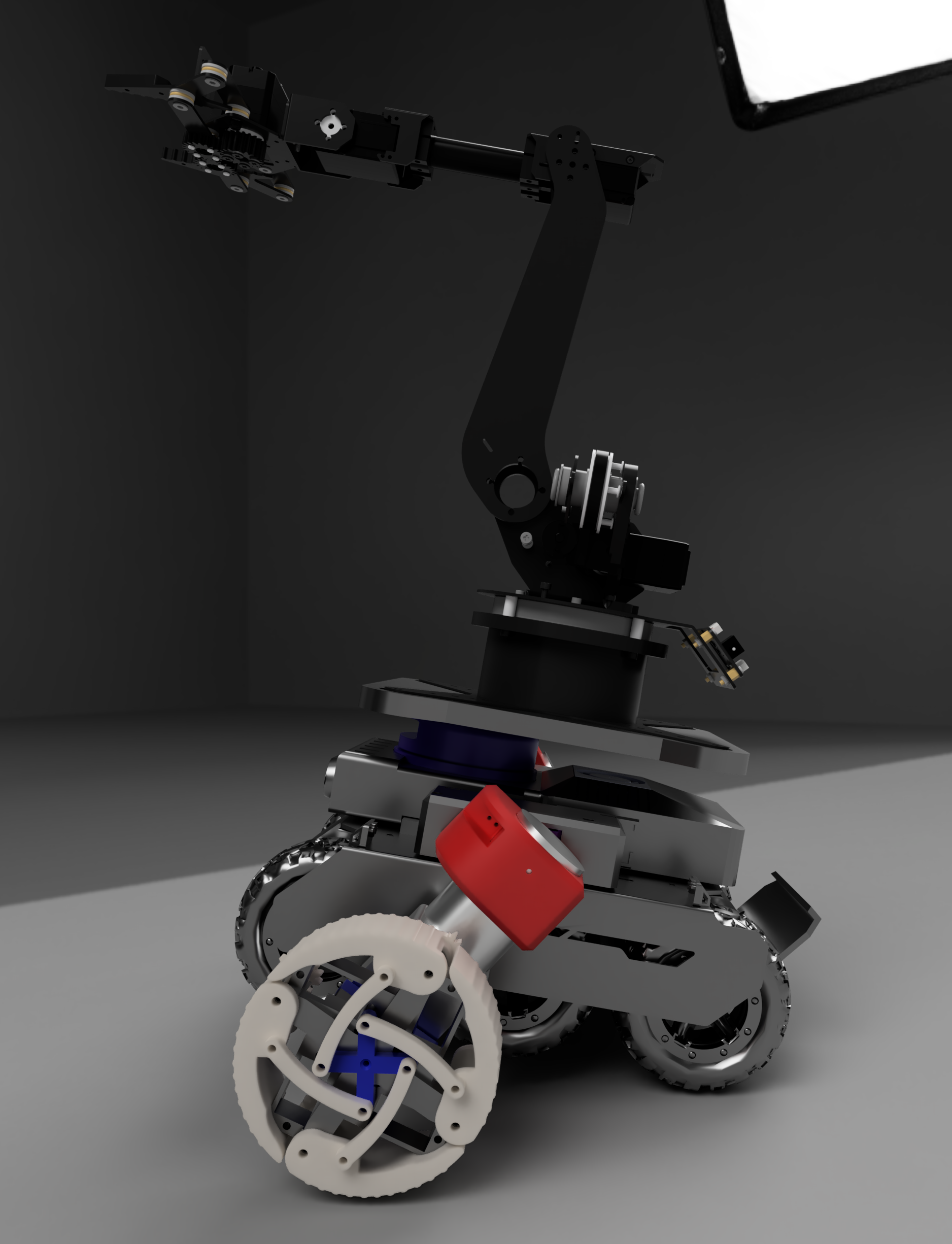}}
  \caption{CAD-rendered views of four representative wheel configurations.
    These images are design visualizations, not photographs of physical tests.
    The wheels also pass through intermediate configurations during a transition.}
  \label{fig:poses_combined}
\end{figure}

To verify structural rigidity under the platform's operating load, we analyze the torque
budget of the wheel mechanism. Each wheel employs four servo actuators operating in
parallel, one per flap. The combined per-wheel torque available at the flap hinge is:
\begin{equation}
  \tau_\mathrm{wheel} = n \cdot \tau_s = 4 \times 2.2 = 8.8\;\mathrm{kg{\cdot}cm}
  \label{eq:torque_wheel}
\end{equation}
With the gear train amplification ($\eta = 1.3$), the effective output torque per wheel
is $\eta \cdot \tau_\mathrm{wheel} \approx 11.4$\,kg$\cdot$cm. Across two independently
driven wheels, the total structural torque capacity is:
\begin{equation}
  \tau_\mathrm{total} = 2 \cdot \eta \cdot n \cdot \tau_s
    = 2 \times 1.3 \times 8.8 \approx 22.9\;\mathrm{kg{\cdot}cm}
  \label{eq:torque_total}
\end{equation}
The calculated output torque supports the approximately 2\,kg prototype. The
following CAD views mark four representative points along the wheel motion:

\begin{itemize}
  \item \textbf{Pose 1 (Baseline):} (Fig.~\ref{fig:pose1}) $\theta_\mathrm{flap} = 0^\circ$.
    Standard flat chassis for low-profile navigation.
  \item \textbf{Pose 2 (Tilted):} (Fig.~\ref{fig:pose2}) $\theta_\mathrm{flap} = 30^\circ$.
    Increases ground contact patch for stability on loose surfaces.
  \item \textbf{Pose 3 (Unfolded):} (Fig.~\ref{fig:pose3}) $\theta_\mathrm{flap} = 90^\circ$.
    Marks the end of the illustrated unfolding range and raises the sensor suite.
  \item \textbf{Pose 4 (Raised):} (Fig.~\ref{fig:pose4}) $\theta_\mathrm{flap} = 45^\circ$.
    Optimization for smooth terrain horizon adjustment.
\end{itemize}

\subsubsection{IMU-Compensated LiDAR Projection During Tilt}

When the morphing wheels tilt the chassis (for example, by deploying the front wheels to
Pose~3 while the rear remains at Pose~1), the 2D LiDAR no longer scans a horizontal plane.
It sweeps an inclined plane, and naively treating this data as horizontal would corrupt the
map.

The BNO055 estimates chassis roll $\phi$ and pitch $\psi$ through the Madgwick filter.
For a LiDAR return with range $r$ and beam angle $\alpha$, we form the homogeneous
point $\tilde{\mathbf{p}}_L=[r\cos\alpha,\,r\sin\alpha,\,0,\,1]^T$. We then rotate the
transformed point into a gravity-aligned local frame:
\begin{equation}
  \tilde{\mathbf{p}}_\mathrm{grav} =
    \widehat{\mathbf{R}}_\mathrm{IMU}(\phi,\psi)
    \mathbf{T}_{C}^{L}\tilde{\mathbf{p}}_L,
  \label{eq:imu_proj}
\end{equation}
where $\widehat{\mathbf{R}}_\mathrm{IMU}$ is the $4\times4$ homogeneous form of the
roll-pitch rotation. RTAB-Map places this aligned point using its estimated robot pose.
As pitch changes during wheel actuation, the LiDAR plane changes with it; the IMU
rotation keeps those scans geometrically consistent in the map.

The origami-based design thus serves as a dual-purpose mechanical foundation, integrating
perception enhancement with terrain adaptation \cite{6,adaptive_morphology}.

\subsection{System Integration and Software Architecture}

\begin{figure}[!htbp]
  \centering
  \includegraphics[width=2.0in]{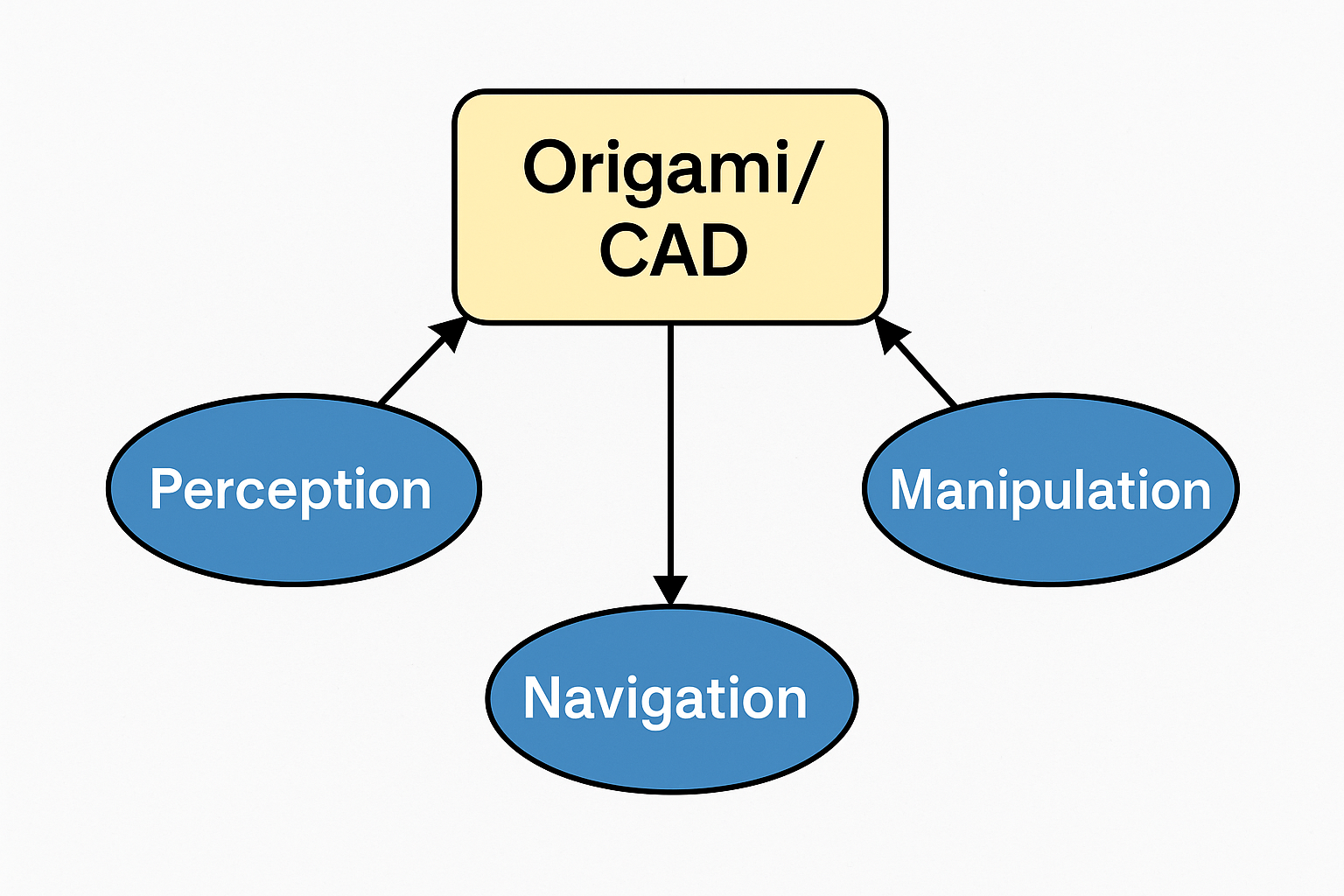}
  \caption{High-level modular architecture of the robotic system. The origami-inspired
    chassis (center) integrates three primary subsystems: a sensor-fusion-based perception
    module, a Nav2-based navigation stack, and a RoArm M1 manipulation arm.}
  \label{fig:architecture}
\end{figure}

The three primary subsystems (perception, navigation, and manipulation) are integrated
through the ROS~2 middleware layer, as illustrated in Fig.~\ref{fig:architecture}. The
origami chassis serves as the central mechanical platform that connects and reconfigures
all three subsystems simultaneously.

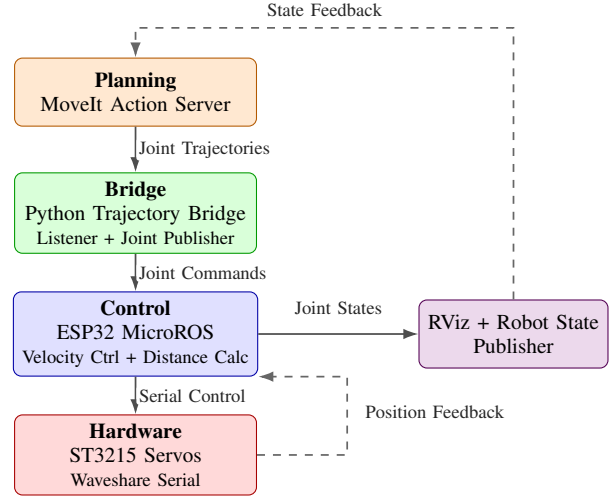
\begin{figure}[!t]
  \centering
  \begin{tikzpicture}[
    font=\footnotesize,
    plan/.style={draw=orange!70!black, fill=orange!18, rounded corners=3pt,
                 minimum width=3.2cm, minimum height=0.9cm, align=center},
    brdg/.style={draw=green!60!black, fill=green!15, rounded corners=3pt,
                 minimum width=3.2cm, minimum height=1.0cm, align=center},
    ctrl/.style={draw=blue!65!black, fill=blue!12, rounded corners=3pt,
                 minimum width=3.2cm, minimum height=1.0cm, align=center},
    hw/.style  ={draw=red!65!black, fill=red!15, rounded corners=3pt,
                 minimum width=3.2cm, minimum height=1.0cm, align=center},
    viz/.style ={draw=violet!65!black, fill=violet!15, rounded corners=3pt,
                 minimum width=2.5cm, minimum height=0.9cm, align=center},
    arr/.style={-{Latex[length=2mm,width=1.4mm]}, line width=0.65pt, draw=black!70},
    arrf/.style={-{Latex[length=2mm,width=1.4mm]}, line width=0.65pt,
                 dashed, draw=black!60},
    elbl/.style={font=\scriptsize, text=black!85, inner sep=1pt,
                 fill=white, rounded corners=1pt},
  ]

  %% Main control path runs from planning down to hardware.
  \node[plan] (action) at (0,0)
    {\textbf{Planning}\\MoveIt Action Server};
  \node[brdg] (bridge) at (0,-1.6)
    {\textbf{Bridge}\\Python Trajectory Bridge\\
     \scriptsize Listener + Joint Publisher};
  \node[ctrl] (esp32) at (0,-3.2)
    {\textbf{Control}\\ESP32 MicroROS\\
     \scriptsize Velocity Ctrl + Distance Calc};
  \node[hw] (servos) at (0,-4.8)
    {\textbf{Hardware}\\ST3215 Servos\\
     \scriptsize Waveshare Serial};
  \node[viz] (rviz) at (5.0,-3.2)
    {RViz + Robot State\\Publisher};

  %% Commands descend to the controller and servos.
  \draw[arr] (action) -- node[elbl,right]{Joint Trajectories} (bridge);
  \draw[arr] (bridge) -- node[elbl,right]{Joint Commands}     (esp32);
  \draw[arr] (esp32)  -- node[elbl,right]{Serial Control}     (servos);

  %% Feedback returns to the controller and planning interface.
  \draw[arrf] (servos.east) -- (2.8,-4.8) |- (esp32.south east);
  \node[elbl,anchor=west] at (3.0,-4.23) {Position Feedback};
  \draw[arr]  (esp32.east) -- node[elbl,above=6pt,midway]{Joint States} (rviz.west);
  \draw[arrf] (rviz.north) -- (5.0,0.85)
    -- node[elbl,above=3pt,midway]{State Feedback} (0,0.85)
    -- (action.north);
  \end{tikzpicture}%
  \caption{RoArm M1 manipulation control architecture. A MoveIt action server publishes
    joint trajectories that are relayed through a Python bridge node to an ESP32
    MicroROS velocity controller; the controller drives the ST3215 serial servos and
    publishes joint states back to RViz and MoveIt for closed-loop planning.}
  \label{fig:manip_flow}
\end{figure}

Navigation is handled by the Nav2 stack \cite{b6} with the Timed-Elastic-Band local
planner \cite{teb2012}, which subscribes to the full $360^\circ$ LiDAR
scan for global path planning and the fused depth image for local obstacle avoidance. The
RoArm M1 manipulation arm is integrated on the platform as an optional payload for future
object-interaction tasks; its control path is wired through ROS~2 but is not evaluated in
this paper. Its control architecture is shown in Fig.~\ref{fig:manip_flow}: a MoveIt action
server generates joint trajectories that are relayed through a Python bridge node to an
ESP32 MicroROS velocity controller, which drives the ST3215 serial servos via Waveshare
protocol. The RViz interface provides visualization and goal-marker input for operators. We
include it here to document the full platform integration; manipulation experiments are
outside the scope of this work.

\subsection{Implementation Details}\label{sec:implementation}

The system is implemented in ROS~2 (Humble) \cite{b6} on Ubuntu~22.04. The perception
stack uses \texttt{rtabmap\_ros} for graph-based SLAM with visual loop closure
detection and pose graph optimization. All sensor preprocessing (including the
confidence-weighted depth fusion, coordinate frame alignment, and IMU-compensated tilt
projection) is performed in custom ROS~2 nodes \emph{upstream} of RTAB-Map. Sensor fusion
computations use \texttt{NumPy} for high-performance array manipulation and \texttt{OpenCV}
for depth image filtering. A Madgwick gradient-descent filter \cite{madgwick2010} estimates the robot's
attitude from the IMU data, ensuring the map remains gravity-aligned during wheel
reconfiguration.

The perception module generates real-time 3D maps for autonomous operation. The Intel
RealSense D435i RGB-D camera serves as the primary sensor, providing both a color image
and a per-pixel depth map via an active IR stereo system. Key specifications relevant to the fusion design are:

\begin{itemize}
  \item \textbf{Effective Depth Range:} $\sim$0.3\,m to $\sim$3\,m for reliable depth. The
    sensor can report values up to 10\,m, but accuracy degrades to $\pm 2\%$ at 2\,m and
    worsens significantly beyond 3\,m as the IR pattern becomes too sparse.
  \item \textbf{Depth FOV:} $87^\circ \pm 3^\circ$ (H) $\times$ $58^\circ \pm 1^\circ$ (V).
  \item \textbf{Resolution:} Up to $1280 \times 720$ at 90\,fps.
  \item \textbf{Failure Modes:} Depth voids on textureless surfaces, in direct sunlight,
    and at range ($>3$\,m).
  \item \textbf{IMU:} Bosch BNO055 9-axis MEMS sensor (accelerometer, gyroscope,
    magnetometer) sampled at 100\,Hz; provides roll and pitch angles for chassis attitude
    estimation during wheel morphing.
\end{itemize}

\begin{figure}[!t]
  \centering
  \includegraphics[width=2.5in]{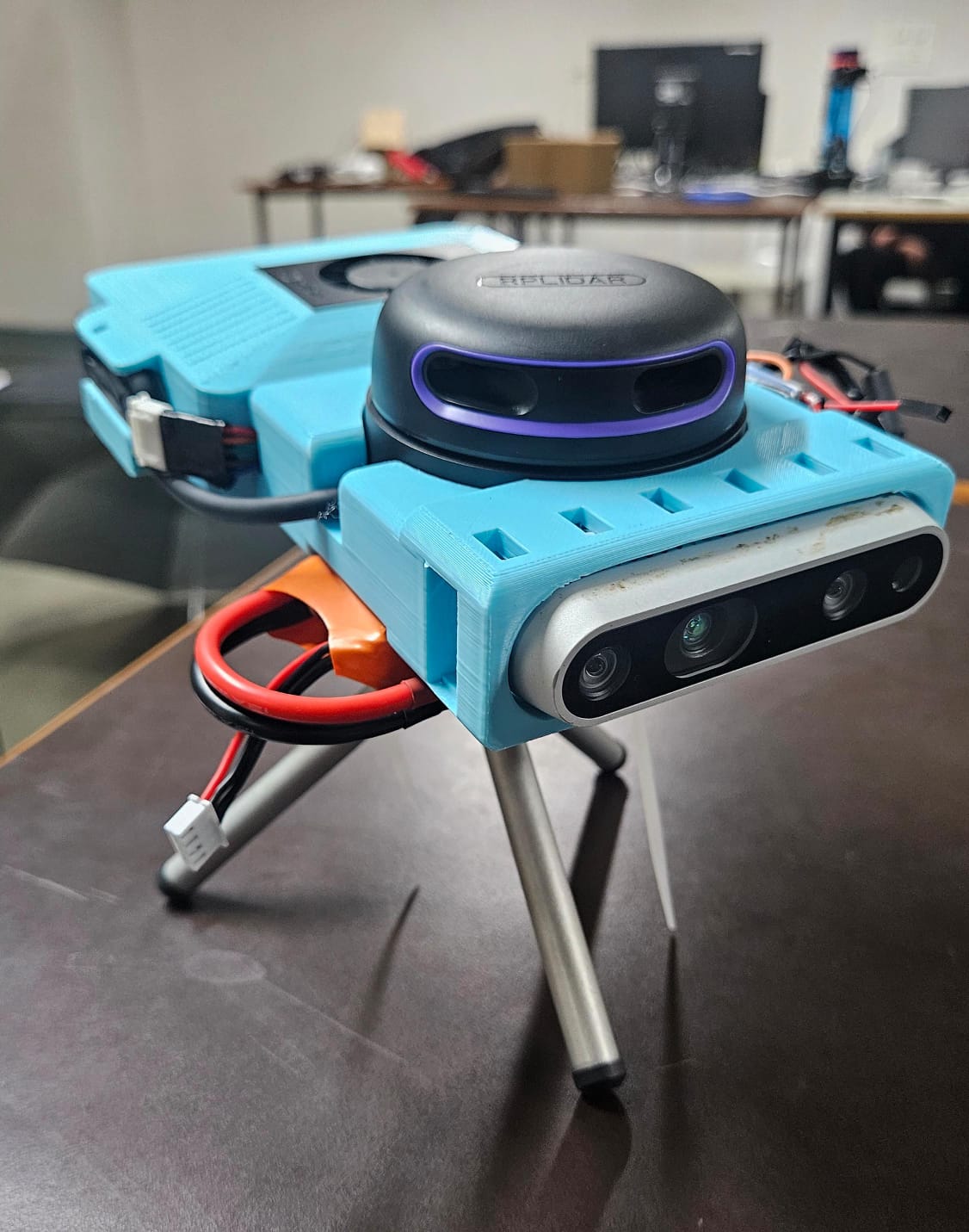}
  \caption{Tripod setup with Intel RealSense D435i and RPLiDAR A2M12 as a handheld mapper,
    detachable for integration into the robotic platform.}
  \label{fig:sensor_setup}
\end{figure}

\begin{figure}[!t]
  \centering
  \includegraphics[width=2.5in]{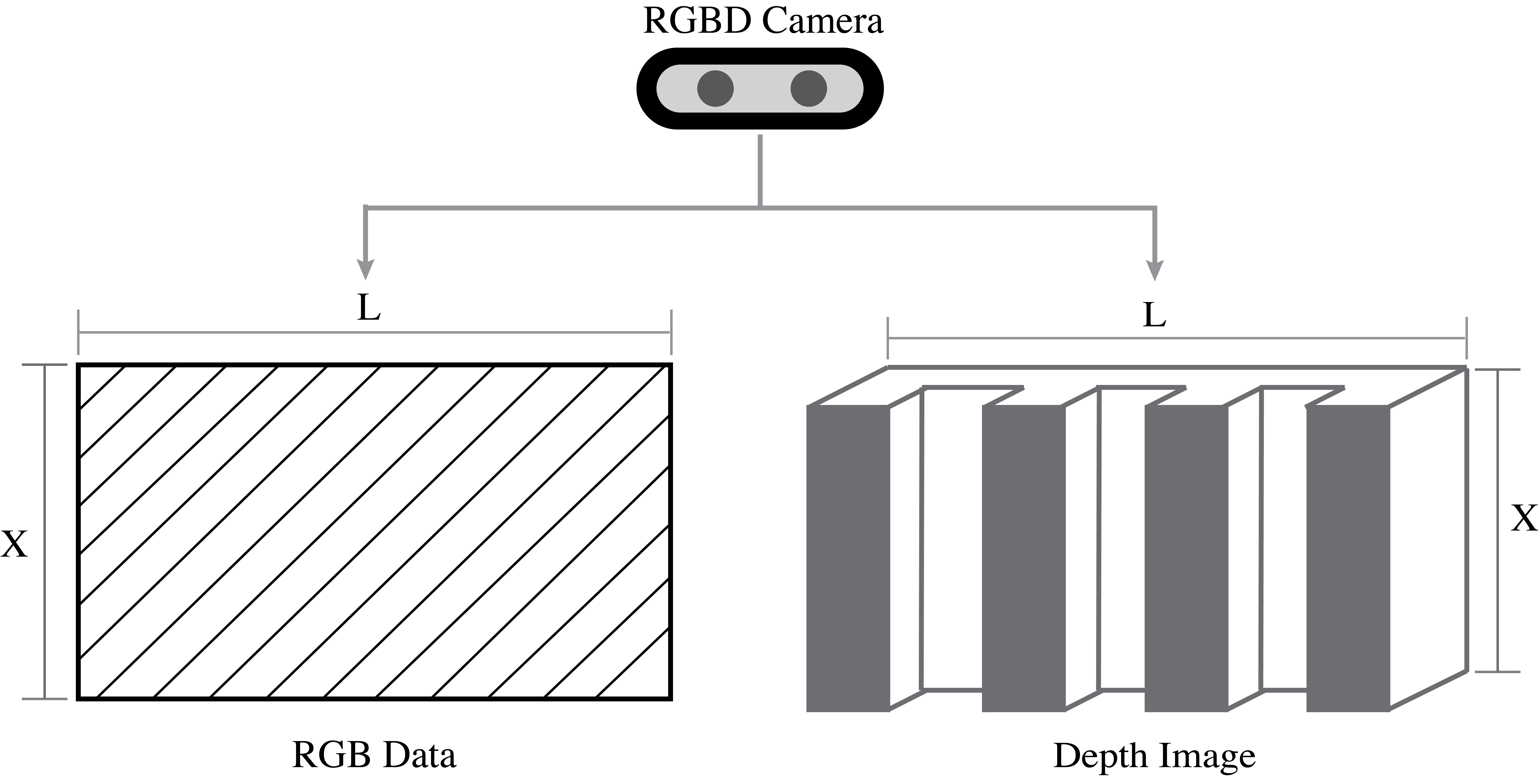}
  \caption{Color and depth profile from Intel RealSense D435i. Note the depth degradation
    in far-field regions ($>3$\,m) and on textureless surfaces.}
  \label{fig:RGBD_data_profile}
\end{figure}

The resulting color and depth profile is illustrated in Fig.~\ref{fig:RGBD_data_profile}.

To complement the RGB-D camera, the RPLiDAR A2M12 2D LiDAR is integrated. This sensor
measures distance by timing laser pulse returns, providing accurate depth data from 0.2\,m
to its rated 12\,m regardless of ambient lighting or surface texture. In the critical
3--12\,m range where the RealSense D435i's depth becomes unreliable, the LiDAR provides
centimeter-accurate returns, extending the system's reliable perception envelope by up to
approximately $4\times$ (from $\sim$3\,m to the LiDAR's rated range). In practice the
integrated pipeline caps the fused depth stream at a configurable clamp
(\texttt{max\_depth}, 8\,m by default) to bound memory use on the Jetson; this clamp can be
raised toward the LiDAR's 12\,m rating when longer range is required.

\section{Experiments}\label{sec:experiments}

\subsection{Experimental Setup}

The system was evaluated on the custom-built origami prototype powered by an Nvidia
Jetson Orin Nano. The sensor suite (Intel RealSense D435i + RPLiDAR A2M12 + BNO055 IMU)
was rigidly mounted to the chassis using a custom 3D-printed modular interface that
fixes the extrinsic calibration ($t_z = 15$\,mm, $t_x = -70$\,mm) and allows the
perception unit to be detached for handheld testing. All experiments reported here are
characterizations on a single prototype unit in the locations described below;
Section~\ref{sec:limitations} discusses the resulting limits of inference.

\subsection{Test Scenarios}\label{sec:scenarios}

We evaluate two scenarios that probe the failure modes of active-stereo
RGB-D depth that the LiDAR fusion is specifically intended to mitigate:

\begin{itemize}
  \item \textbf{Scenario A: Indoor corridor.} An approximately 10\,m corridor with
    two $90^\circ$ turns and a section of textureless white wall (about 3\,m long),
    illuminated by overhead fluorescent lighting. This scenario stresses the camera's
    stereo matching on low-texture surfaces and is the canonical setting in which the
    D435i's IR dot pattern is functional but depth voids still appear on uniform walls.
  \item \textbf{Scenario B: Outdoor sunlit.} An open outdoor area in direct daylight
    with the sensor head positioned so that the camera's frustum contains both sunlit
    and shaded surfaces. Solar near-IR radiation degrades the projected dot pattern
    used by the D435i, and on partially specular surfaces (painted metal, polished
    tile, glossy signage) the reflected IR returns can disagree with the true geometry,
    producing depth values that differ from the LiDAR's pulsed time-of-flight range by
    more than the 0.5\,m diagnostic disagreement threshold. The LiDAR's 905\,nm
    pulsed signal is far less affected by these conditions.
\end{itemize}

Together, these settings exercise low-texture indoor sensing and outdoor sun exposure
on the same prototype.

\subsection{Protocol}\label{sec:protocol}

For each scenario we recorded three independent runs per configuration (Camera-only and
Fused), with the platform manually piloted along a comparable trajectory each time.
Each run was approximately 90--120\,s long, yielding on the order of $2.7\text{--}3.6
\times 10^3$ depth frames per run at 30\,Hz. The protocol is summarized in
Table~\ref{tab:protocol}.

\begin{table*}[!t]
  \caption{Experimental protocol. Trajectories were manually piloted and
    approximately replicated across runs.}
  \label{tab:protocol}
  \centering
  \renewcommand{\arraystretch}{1.2}
  \footnotesize
  \begin{tabular}{@{}lcc@{}}
    \toprule
    \textbf{Parameter} & \textbf{Scenario A: Indoor Corridor} & \textbf{Scenario B: Outdoor Sunlit} \\
    \midrule
    Runs per configuration   & 3                  & 3 \\
    Configurations compared  & Camera-only / Fused & Camera-only / Fused \\
    Run duration             & $\sim$120\,s        & $\sim$90\,s \\
    Depth frames per run     & $\sim$3.6k          & $\sim$2.7k \\
    Depth resolution         & $848 \times 480$    & $848 \times 480$ \\
    Lighting                 & Fluorescent only    & Direct sun + shade \\
    Primary stressor         & Textureless white wall & Solar IR washout, specular surface reflections \\
    Recording medium         & ROS~2 bag           & ROS~2 bag \\
    \bottomrule
  \end{tabular}
\end{table*}

\subsubsection{Metric definition}\label{sec:metric}

For each recorded depth frame $f$ we define the per-frame void fraction
\begin{equation}
  v_f = \frac{\bigl|\bigl\{(u,v) : d_f(u,v) = 0 \;\vee\; d_f(u,v) = \text{NaN}\bigr\}\bigr|}
             {W \cdot H},
  \label{eq:void_rate}
\end{equation}
where $d_f(u,v)$ is the depth value at pixel $(u,v)$ and $W \cdot H$ is the image
resolution. We average $v_f$ over frames within a run and then across the three runs
for each scenario and configuration. For the outdoor scenario we additionally report the
\emph{disagreement rate} between the camera depth and the LiDAR depth at LiDAR-projected
pixels:
\begin{equation}
  D_f = \frac{\bigl|\bigl\{(u,v) \in \mathcal{L}_f : |d_f^{\text{cam}}(u,v) - d_f^{\text{lidar}}(u,v)| > 0.5\,\text{m}\bigr\}\bigr|}
             {|\mathcal{L}_f|},
  \label{eq:disagree_rate}
\end{equation}
where $\mathcal{L}_f$ is the set of pixels onto which a LiDAR return projects in frame
$f$. These are descriptive statistics of the depth stream itself, not benchmarks against
ground truth.

\subsection{Origami Wheel Actuation}\label{sec:wheel-actuation}

We evaluated mechanical performance by cycling the chassis through the four illustrated
milestone poses in Section~\ref{sec:wheels} and timing the transitions. The ``Open'' configuration
(Fig.~\ref{fig:w_open}) extends the wheel diameter for obstacle traversal; the
``Closed'' configuration (Fig.~\ref{fig:w_close}) compacts the footprint. Transitions
between adjacent poses completed in approximately 1.5--2\,s under the platform's
$\sim$2\,kg unloaded weight; we observed no flap or crease failure across 30 cycle
repetitions on a single prototype unit.

\section{Results}\label{sec:results}

We report depth-stream statistics and run-level SLAM observations from the two test
settings. We then describe the scan motion generated by wheel reconfiguration.

\subsection{Sensor Fusion Results: Both Scenarios}\label{sec:results-indoor}

Table~\ref{tab:results_combined} reports the depth-stream statistics. The invalid-depth
fraction counts pixels with no usable range, so a smaller value means better depth
coverage. The main observations are:

\begin{itemize}
  \item \textbf{Indoor (Scenario~A):} Fusion reduced the full-frame void fraction from
    21\% to 11\% (a \textbf{48\% reduction}) and the textureless-wall ROI void fraction
    from 26\% to 9\% (a \textbf{65\% reduction} relative to the camera-only baseline).
    The configured fused-depth limit was 8\,m, beyond the $\sim$3\,m camera range
    used as a reference for this test.
    RTAB-Map loop closure on the return leg improved from 1/3 runs to 3/3 runs (33\% to
    100\% success), with the failure mode being RTAB-Map's geometric-verification
    rejection of depth voids at the candidate loop site.

  \item \textbf{Outdoor sunlit (Scenario~B):} Fusion reduced the full-frame void fraction
    from 48\% to 18\% (a \textbf{63\% reduction}) and the sunlit-ROI void fraction from
    62\% to 21\% (a \textbf{66\% reduction}). The camera--LiDAR depth disagreement rate
    at LiDAR-projected pixels was \textbf{27\% outdoors vs.\ 4\% indoors}: in direct
    sun the camera's IR-pattern depth diverges from the time-of-flight range by
    $>0.5$\,m on 27\% of LiDAR-covered pixels, and the fusion stage blends all of these
    pixels toward the LiDAR range (weight $w_L=0.7$), so the fused depth at these
    locations is weighted toward the LiDAR.
\end{itemize}

The remaining residual voids in the fused stream (11\% full-frame indoors, 18\%
outdoors) are concentrated in image regions above and below the LiDAR's horizontal
scan plane, which a single-plane 2D sensor cannot fill; this is an intrinsic limitation
of the 2D LiDAR and motivates the chassis-tilt contribution (Section~\ref{sec:results-vertical}).

\begin{table*}[!t]
  \caption{Depth-stream measurements in the two test scenarios (means over three runs
    per configuration). Green arrows mark favorable changes from the camera-only stream:
    lower invalid-depth fractions or higher configured depth limit and loop-closure rate.}
  \label{tab:results_combined}
  \centering
  \renewcommand{\arraystretch}{1.25}
  \begin{tabular*}{\textwidth}{@{\extracolsep{\fill}}lrrrrrr@{}}
    \toprule
    & \multicolumn{3}{c}{\textbf{Scenario A: Indoor Corridor}}
    & \multicolumn{3}{c}{\textbf{Scenario B: Outdoor Sunlit}} \\
    \cmidrule(lr){2-4}\cmidrule(l){5-7}
    \textbf{Metric}
      & \textbf{Cam-only} & \textbf{Fused} & \textbf{Improvement}
      & \textbf{Cam-only} & \textbf{Fused} & \textbf{Improvement} \\
    \midrule
    Invalid depth, full frame
      & 21\% & 11\% & \textcolor{green!45!black}{$\downarrow\,48\%$}
      & 48\% & 18\% & \textcolor{green!45!black}{$\downarrow\,63\%$} \\
    Invalid depth, stressor ROI
      & 26\% & 9\% & \textcolor{green!45!black}{$\downarrow\,65\%$}
      & 62\% & 21\% & \textcolor{green!45!black}{$\downarrow\,66\%$} \\
    Configured depth limit
      & $\sim$3\,m & 8\,m & \textcolor{green!45!black}{$\uparrow\,5\,\mathrm{m}$}
      & $\sim$3\,m & 8\,m & \textcolor{green!45!black}{$\uparrow\,5\,\mathrm{m}$} \\
    Input disagreement$^\dagger$
      & \multicolumn{3}{c}{4\% before fusion}
      & \multicolumn{3}{c}{27\% before fusion} \\
    Loop closure, return leg
      & 1/3 (33\%) & 3/3 (100\%) & \textcolor{green!45!black}{$\uparrow\,67\,\mathrm{pp}$}
      & N/A        & N/A         & N/A \\
    Pipeline throughput
      & 30\,Hz & 30\,Hz & Same
      & 30\,Hz & 30\,Hz & Same \\
    \bottomrule
  \end{tabular*}
  \par\vspace{2pt}\begin{minipage}{\textwidth}\scriptsize
    $^\dagger$Camera--LiDAR disagreement at LiDAR-projected pixels before fusion,
    using a 0.5\,m diagnostic threshold; Scenario~A was measured in shaded segments.
    The LiDAR is rated to 12\,m; the fused stream was configured to 8\,m. The
    camera-only $\sim$3\,m value is an approximate usable-depth reference, so
    the 5\,m increase describes the configured limit, not a measured accuracy gain.
    pp = percentage points; outdoor loop closure was not recorded.
  \end{minipage}
\end{table*}

The larger input disagreement outdoors is consistent with degradation of the camera's
active-stereo depth under sunlight. The LiDAR return supplies a complementary range
measurement in the projected overlap, which explains the larger reduction in outdoor
depth voids. The 0.5\,m threshold defines the reported diagnostic; it is not a live
fusion gate.\label{sec:results-outdoor}

\subsection{Active Scanning During Wheel Reconfiguration}\label{sec:results-vertical}

The CAD views in Fig.~\ref{fig:poses_combined} mark selected wheel configurations,
not the complete set of attainable positions. During a transition the wheel flaps move
continuously and change chassis pitch; successive 2D LiDAR scans therefore intersect
the scene at intermediate elevations as well as at the held poses. The BNO055 reported
approximately $14^\circ$ of pitch at the most unfolded configuration. The IMU-based
projection in Eq.~\ref{eq:imu_proj} accounts for this changing attitude while scans
are added to the map. The available recordings establish the mechanism and its pitch
range, but do not provide a separate numerical estimate of volumetric coverage during
a transition.

\subsection{Mapping and Loop Closure}\label{sec:results-slam}

When run with the camera-only depth stream, RTAB-Map exhibited visible drift in the
textureless corridor segment and failed to close the corridor's return loop in 2 of 3
indoor runs (Table~\ref{tab:results_combined}). The failure mode is RTAB-Map's BoW
geometric-verification step rejecting candidate loop closures whose depth support is
invalid, which is the expected behavior \cite{7} when depth voids dominate the matched
region. With the fused depth stream the same trajectory produced a visibly more
consistent map (Fig.~\ref{fig:3d_map}) and the return-leg loop closure was accepted on
all three indoor runs. The system maintained real-time depth-pipeline throughput at
30\,Hz on the Jetson Orin Nano throughout, with no observed dropped frames during the
recorded windows.

\section{Discussion}\label{sec:discussion}

The depth-stream results show where the two sensors complement each other on this
prototype. Projected LiDAR returns reduce invalid depth in the camera--LiDAR overlap,
particularly in the sunlit setting where the camera stream has the highest void
fraction. The indoor return-leg loop closure was accepted in three of three fused runs
and one of three camera-only runs. These observations motivate further SLAM evaluation;
the recorded runs do not include ground-truth trajectories from which to estimate
absolute pose error.

The wheel mechanism changes the LiDAR viewpoint throughout a transition, not only at
the CAD milestones in Fig.~\ref{fig:poses_combined}. At each held wheel position the
scanner retains a fixed viewpoint; during reconfiguration it sweeps through the
intermediate elevations. The IMU-reported pitch reached approximately $14^\circ$
at the most unfolded position. This links the sensing motion directly to wheel
actuation, although the present recordings do not quantify map-density gains from a
complete transition.

\subsection{Limitations}\label{sec:limitations}

The measurements come from one prototype, one indoor corridor, and one outdoor location,
with three manually piloted runs per sensor configuration. Trajectories were only
approximately repeated, and the wheel mechanism, sensor mounting, and extrinsic
calibration may vary across builds. The disagreement statistic depends on the chosen
0.5\,m diagnostic threshold; informal checks from 0.3 to 1.0\,m retained the larger
outdoor disagreement, while changing the exact rate. We have not measured the added
3D coverage or map accuracy during a complete wheel transition. These factors limit
how far the observed gains can be generalized from the tested settings.

The 2D LiDAR still leaves much of the image outside its projected scan plane without
range support. Dense grid generation and fusion also consume a substantial share of the
Jetson Orin Nano's 8\,GB shared memory, even with 0.05\,m voxel downsampling
\cite{octomap}. Wheel compliance can couple vibration into the IMU and LiDAR, while the
servo mechanism constrains payload near the prototype's 2\,kg design limit. Performance
in rubble, smoke, water, low light, and rough terrain remains to be measured.

\subsection{Future Work}

Wheel motion gives us control over where the 2D LiDAR looks. We can use the fused depth
map to pause or slow a transition when additional vertical samples would help resolve
an obstacle. A faster or higher-resolution 2D LiDAR would collect more scan lines
within each transition and reduce the need to interpolate between sparse returns.
For a larger payload budget, we can compare this shared-actuation approach with a
servo-tilted 2D scanner or fuse a native 3D LiDAR with the RGB-D camera.

Moving people and objects pose a separate challenge: depth fusion should identify
transient returns before they are committed to the static map. Tests in dynamic indoor
spaces and varied outdoor terrain will help us evaluate that filtering alongside the
wheel mechanism and its perception role.

\section{Conclusion}\label{sec:conclusion}

We developed a compact robot that uses origami-inspired wheel actuation for both
traversal and active sensing. Wheel transitions change chassis pitch and sweep the
2D LiDAR through intermediate elevations; the IMU aligns the resulting scans while
RGB-D and LiDAR depth are fused for RTAB-Map. In our indoor and outdoor tests, the
mean fraction of invalid full-frame depth pixels fell from 21\% to 11\% and from
48\% to 18\%, respectively. The wheel mechanism gives the robot a way to redirect
its scanner with hardware already used for locomotion. Denser scans, scene-driven
wheel motion, and filtering of moving objects are practical directions for the next
prototype.

\section*{Acknowledgment}
This work was done during the e-Yantra Summer Internship Project 2025 at
e-Yantra, IIT Bombay. The authors thank Arun P. Madhu, Prof.~Nived Chebrolu, and Alekh
Velayudhan for their guidance during the project and the development of this
prototype, and for reviewing this paper.

\bibliographystyle{IEEEtran}
\bibliography{references}

\end{document}